\documentclass{article}

\usepackage{iclr2027_conference,times}
\usepackage[utf8]{inputenc}
\usepackage[T1]{fontenc}
\usepackage{amsmath,amssymb,amsthm}
\usepackage{booktabs}
\usepackage{graphicx}
\usepackage{algorithm}
\usepackage{algorithmic}
\usepackage{tabularx}
\usepackage{longtable}
\usepackage[hidelinks]{hyperref}
\usepackage[expansion=false]{microtype}

\newcommand{\acc}{\operatorname{Acc}}
\newcommand{\macc}{\operatorname{mAcc}}
\newcommand{\iou}{\operatorname{IoU}}

\title{Where Grounding Accuracy Lives on the IoU Curve: Label-Free Inference-Time Boundary Refinement}
\iclrfinalcopy
\author{Bo Ma \\ Auckland University of Technology}

\begin{document}

\maketitle
\lhead{Under review as a conference paper at ICLR 2027}

\begin{abstract}
Vision--language models can identify the correct referent while returning an
imprecise bounding box.  We study whether a frozen direct-answer model can
use its own prediction to allocate one additional localized observation
without accessing target annotations at inference.  Label-free precision
refinement (LFPR) routes predicted-small regions to a higher-resolution
pass, re-grounds the expression inside a context crop, admits a candidate
only under fixed geometric guards, and returns a fixed coordinate-wise
midpoint.  We report results across three evidence tiers.  On 31,921
retrospective Ref-L4 expressions, LFPR raises mAcc$_{0.5:0.95}$ from
72.947\% to 76.013\% (Acc@0.5 88.531\%$\to$89.725\%, Acc@0.9
55.788\%$\to$61.142\%).  A frozen transfer to 30,969
RefCOCO/RefCOCO+/RefCOCOg expressions improves every dataset at Acc@0.5,
mAcc, and mean IoU (pooled mAcc $+0.645$, Acc@0.5 $+0.817$), while Acc@0.9
is unchanged overall: routing alone gains $+1.162$ points there, but crop,
guards, and fusion give back $-1.192$, offsetting rather than showing no
strict-IoU effect.  A prospective, image-disjoint Flickr30K Entities
evaluation improves every endpoint (mAcc $+0.973$, Acc@0.9 $+1.022$), more
strongly under a single-box variant (mAcc $+2.575$, Acc@0.9 $+3.689$).  The
same operator applied to two released grounding specialists improves every
endpoint (Acc@0.9 $+1.569$/$+6.716$ for EGM-4B/8B) at roughly twice the
latency, composing with specialist training rather than replacing it.  A
genuine unguarded control (guard removed from the same candidates)
underperforms the incumbent on every metric, showing the guard is
load-bearing.  Together, these results show that referent selection and
boundary precision are partially separable, with different components
moving opposing regions of the IoU curve -- behavior a single threshold
cannot reveal.
\end{abstract}

\section{Introduction}
\label{sec:introduction}

Referring expression grounding asks a vision-language model (VLM) to map a
natural-language description to the image region it denotes.  Modern VLMs are
often strong enough to identify the correct object while still returning a box
that is too coarse, especially for long expressions, small objects, and images
whose native resolution falls below the model's effective visual input floor.
The distinction matters because a small coordinate error can leave an Acc@0.5
outcome unchanged while causing a failure at stricter thresholds: a method
improving only headline Acc@0.5 may hide a useful localization change, and one
improving only strict thresholds should not be sold as a general accuracy
gain.

The usual responses -- train a grounding head, add a detector, or spend more
computation on every image -- each change the comparison or incur a
substantial deployment cost.  We study a narrower question: can a frozen
direct-answer VLM use its own first prediction to decide when a second,
localized look is worth the cost?  The question is deliberately operational,
since the routing signal must be available before the ground-truth box is
known.  We therefore stratify our evidence into three tiers: a retrospective
Ref-L4 analysis whose published test distribution informed method
development, a frozen cross-dataset transfer to the RefCOCO family that
retuned nothing, and an image-disjoint Flickr30K confirmation whose images
and annotations played no role in method development.

Our procedure, label-free precision refinement (LFPR), has two components.
First, it applies pixel-area size buckets to the incumbent's \emph{predicted}
box; predictions below the small-object threshold are re-evaluated at a higher
image-pixel budget, and all other rows keep the default-resolution answer.
Second, the resulting incumbent is re-grounded inside a context-expanded crop,
where a candidate is admitted only if zoom-consistency and geometric guards
pass, and the final box is the coordinate-wise midpoint of incumbent and
candidate.  That midpoint is a fixed convex combination chosen before the
reported run: it lets the crop adjust boundaries without letting one noisy
second answer replace a strong first one.  We do not claim it is an optimal
shrinkage estimator.

The main result is positive but bounded.  On the analyzed Ref-L4 test split
LFPR improves the frozen Qwen3-VL-8B~\citep{qwen3vl2025} baseline from 88.531\% to 89.725\% at
Acc@0.5, with the largest differences at Acc@0.75 and Acc@0.9.  Because the
same published test distribution informed diagnostics and method development,
we treat that estimate and its interval as retrospective; it now exceeds a
cross-protocol 89.5\% comparison point, but the comparison remains
cross-protocol and parameter-asymmetric (Section~\ref{sec:sota}), so we
still make neither a confirmatory nor an absolute state-of-the-art claim.

This paper makes three contributions within that evidence boundary:
\begin{enumerate}
  \item We formulate a label-free, inference-only precision-refinement
        pipeline whose accept/reject decision uses only geometry computed
        from the model's own outputs, and show it improves a frozen
        generalist, two released specialists, and an image-disjoint
        prospective benchmark without any weight update.
  \item We show that referent selection and boundary placement are separable
        capabilities that trade off along the IoU threshold axis: among
        released specialists on identical rows, the best-Acc@0.5 system has
        the worst Acc@0.9 and mean accuracy, so a permissive-threshold
        ranking can invert the localization-quality ordering.
  \item We report a same-row component decomposition: the resolution floor
        and crop, with the guard active, each buy strict-IoU accuracy, while
        the guards and midpoint trade some of it back for the largest
        Acc@0.5 gain.  A genuine unguarded ablation of the same candidates
        does not recover that accuracy, confirming the guard is
        load-bearing rather than a discarded trade-off.
\end{enumerate}

\section{Related Work}
\label{sec:related}

Referring-expression benchmarks such as RefCOCO and RefCOCOg established the
standard single-region grounding setting
\citep{yu2016refcoco,mao2016refcocog}; Ref-L4 revisits it for contemporary
multimodal models with longer expressions and a stricter re-annotated
protocol \citep{chen2025refl4}.  We use the official Ref-L4 evaluator and
report the complete threshold profile rather than only Acc@0.5.

Supervised REC ranges from end-to-end text-conditioned detection
\citep{kamath2021mdetr} to large-scale grounded pretraining
\citep{li2022glip}, with recent work on dynamic visual routing
\citep{chen2025dvin} and zero-shot REC via debiased attention and
programmatic relation reasoning \citep{chen2025zeroshot}.  Language-guided
open-set detectors \citep{liu2023groundingdino,minderer2023owlv2} offer
another route, but adopting one would add a second model, a proposal cache,
and a selection policy, changing the estimand from ``is the incumbent's own
answer precise enough'' to ``can an external candidate be selected
correctly.''  All of these change supervision or add machinery; LFPR keeps
one direct-answer checkpoint frozen and changes only its inference path.

LFPR is structurally a coarse-to-fine cascade \citep{dai2025c3vg}, but every
stage reuses the same frozen weights and the consistency between stages is
enforced by fixed midpoint fusion and geometric guards rather than a
learned loss (a directly trained adaptation did not improve on the frozen
baseline, Section~\ref{sec:negative-controls}).

The closest family is test-time self-correction.  Iterative loops that ask a
model to revise its own box can show gains that depend on an oracle stopping
rule and shrink or reverse without ground-truth access -- the
``self-correction mirage'' \citep{tripathy2026itervisual}; related work
scales test-time perception through explicit zoom \citep{jiang2026testtimeperception}
or an entropy-guided crop-and-reground loop \citep{gropl2026entropygradient}.
The closest training-free method on the same benchmark family is
Chain-of-Caption, which supplies additional visual and textual context
through tool use \citep{pang2026chainofcaption}; it improves context rather
than the box itself, so it is complementary rather than a substitute for a
geometry-based accept/reject step, and we did not run a head-to-head
comparison since its released code targets a different backbone family --
the same compatibility check applies to EGG and TTSP above, expanded in
Appendix~\ref{app:baseline-incompatibility}.
LFPR differs in three respects: at most one crop attempt rather than a loop,
an accept/reject decision computed only from the geometry of the two
candidate boxes, and a fixed conservative midpoint rather than outright
replacement.  A matched zero-shot self-correction control on the same
incumbent regresses the headline metric
(Appendix~\ref{app:relocated}), consistent with the mirage finding and
motivating the guards.  A same-backbone mechanism control substitutes a
frozen, label-free confidence threshold for the geometry guard on the same
admission point -- not a reproduction of EGG or TTSP, whose released code
targets a different backbone -- and the confidence-gated arm does not
recover even the guard's modest Acc@0.5 gain over routing
(Appendix~\ref{app:controls}), evidence for the accept/reject step's
specifically geometric criterion rather than any label-free filter.

Efficient Grounding Models (EGM) instead scale generated reasoning tokens for
small VLMs and report strong accuracy--latency results on the same backbone
family \citep{zhan2026egm}; they are our released-specialist comparison in
Section~\ref{sec:specialists}.

\section{Problem Formulation and Method}
\label{sec:method}

\subsection{Direct-answer incumbent}

Let $I$ be an image and $q$ its referring expression.  A frozen VLM
$f_{\theta}$ returns a normalized box
$b_0=f_{\theta}(I,q)\in[0,1]^4$, or an invalid output that is retained as an
incumbent miss.  For a ground-truth box $y$, the threshold utility is

\begin{equation}
  u_t(b,y)=\mathbb{1}\{\iou(b,y)>t\}.
  \label{eq:utility}
\end{equation}

Averaged over a population, $A(t)=\mathbb{E}[u_t]=\Pr(\iou(b,y)>t)$ is the
survival function of the IoU distribution, so $\mathbb{E}[\iou(b,y)]=\int_0^1
A(t)\,dt$: an intervention can raise $A(t)$ at one threshold while lowering it
at another, which is the lens we use throughout Section~\ref{sec:results} and
Section~\ref{sec:transfer} rather than reporting a single operating point.
The strict inequality is the official Ref-L4 convention.  We use
$\macc_{0.5:0.95}$, the mean of $100\,\mathbb{E}[u_t]$ over
$t\in\{0.50,0.55,\ldots,0.95\}$, as the manuscript's primary summary because
it captures the reported IoU profile.  This designation is retrospective and
does not confer confirmatory status.  We report
$\acc_{0.5}=100\,\mathbb{E}[u_{0.5}]$ as the standard benchmark operating
point, with $\acc_{0.75}$ and $\acc_{0.9}$ as strict-IoU diagnostics.

\subsection{Label-free resolution routing}

Given an incumbent box and the original image dimensions $(W,H)$, convert
the box to pixel coordinates and define

\begin{equation}
  s(b_0)=\sqrt{\max(0,x_2-x_1)\max(0,y_2-y_1)}.
  \label{eq:size}
\end{equation}

The frozen bucket function is small if $s<128$ pixels, medium if
$128\leq s\leq256$, and large otherwise.  LFPR routes only the small
bucket to a second full-image inference with the already validated
minimum-pixel budget of 4,194,304.  Let $b_0'$ denote this second answer for
flagged rows and $b_0'=b_0$ otherwise.  Crucially, the bucket is computed
from $b_0$, not $y$; the ground-truth box is used only for evaluation.

This router is a label-free proxy for a resolution deficit, not an estimate
of correctness: it can route a wrong large prediction or miss a small target
hidden by a large incumbent box.  The predicted-box router agrees with the
annotation-derived bucket on 88.85\% of pilot rows and 89.83\% of official
rows -- a reasonable, though imperfect, substitute for the oracle diagnostic
it is designed to approximate without reading a label.

The routing decision raises the pixel floor rather than lowering a ceiling:
under the default image policy, essentially none of the small-bucket images
already exceed the higher pixel budget used for the second pass, so an
upper-limit policy would be mechanically inert for exactly the rows the
router targets, and raising the minimum pixel budget is the only lever that
changes the visual evidence available to the model for these regions.

\subsection{Local re-grounding and conservative fusion}

For $b_0'=(x_1,y_1,x_2,y_2)$, let $c_x=(x_1+x_2)/2$ and
$c_y=(y_1+y_2)/2$.  The context crop $C_\gamma(b_0')$ has width
$\min\{1,\gamma(x_2-x_1)\}$ and height
$\min\{1,\gamma(y_2-y_1)\}$ with $\gamma=2.0$; its center is shifted only
as needed to keep the window inside $[0,1]^2$.  Pixel crop boundaries use
floor for the upper-left and ceiling for the lower-right before clipping to
the original image.  The frozen VLM is asked to locate the expression inside
this crop and emit crop-normalized coordinates.  The candidate is mapped
affinely back to the original image, producing $b_c$.

Let $a$ be the center of $b_0'$ after mapping the incumbent into crop
coordinates, and let $c(b_c^{C})$ be the center of the crop-coordinate
candidate.  The zoom-consistency distance is

\begin{equation}
 z(b_c,b_0')=\frac{\lVert c(b_c^{C})-a\rVert_2}{\sqrt{1/2}}.
 \label{eq:zoom}
\end{equation}

The crop candidate is accepted only if all three guards pass:

\begin{align}
  z(b_c,b_0') &\leq 0.35,\\
  \iou(b_c,b_0') &\geq 0.25,\\
  0.5 &\leq A(b_c)/A(b_0')\leq2.0,
  \label{eq:guards}
\end{align}

where $A$ is box area.  If a guard or parser check fails, $b_c$ is discarded
and the pre-crop box is retained; an invalid first pass is retained as a
miss.  No coordinate is synthesized and no row is removed at any stage.  The
final prediction is

\begin{equation}
  \widehat b=\begin{cases}
  \frac{1}{2}b_0'+\frac{1}{2}b_c,&\text{if all guards pass},\\
  b_0',&\text{otherwise}.
  \end{cases}
  \label{eq:fusion}
\end{equation}

No score or target box is used in the routing or fusion decision, and every
model output is parsed and canonicalized before use: coordinates must be
finite, ordered, inside $[0,1]^2$, and define positive area.
Appendix~\ref{app:repro} gives the exact prompts, pixel budgets, and clipping
conventions.  Historical records state that the fusion
weight, crop context, resolution setting, and guards were fixed before the
final full-split run, but the broader data-access chronology is not
independently auditable, so Section~\ref{sec:protocol} treats all intervals as
retrospective.

\subsection{Complete procedure}

Appendix~\ref{app:algorithm} states the full label-free pipeline in pseudocode.  Every
decision after the first incumbent prediction is a deterministic function
of already-observed model outputs and image geometry; no step consults the
ground-truth box $y$, which appears only in the evaluator.  The procedure
performs at most one additional full-image inference (for predicted-small
rows) and exactly one crop inference for every valid incumbent, so its cost
is two forward passes for the roughly four in five rows the router does not
flag and three for the remainder.  Only an invalid first-pass output (an
incumbent miss) short-circuits the procedure to a single forward pass.

Figure~\ref{fig:pipeline} (appendix) shows the same procedure as a data-flow
diagram: every row reaches the crop-and-guard stage, but only predicted-small
rows do so after the extra high-resolution pass.

\section{Experimental Protocol}
\label{sec:protocol}

\subsection{Benchmark, model, and splits}

We use the complete Ref-L4 test split: 31,921 expressions over 9,467 unique
images, with COCO and Objects365~\citep{shao2019objects365} source labels retained for stratified
analysis.  The baseline and all treatment predictions use the same frozen
Qwen3-VL-8B checkpoint, prompt family, parser, coordinate convention, and
official evaluator.  The method is not trained on Ref-L4.

The shared 2,000-row crop/refinement pilot (1,713 images) was deterministically
selected from the published Ref-L4 pool after excluding images in named
earlier pilots, then reused first for conservative crop fusion and then for
complete LFPR, so it is not an untouched confirmation set.  Annotation-size
and source diagnostics on the complete published test distribution also
influenced the research program before the final 31,921-row run; ``one read
per method'' does not undo this program-level adaptation.  We accordingly
call both the pilot and full-split results retrospective and use the
image-cluster bootstrap to describe sampling variability under the frozen
comparison, not a confirmatory error rate (Appendix~\ref{app:data-governance}
records what can and cannot be reconstructed from the archived manifests).

\subsection{Implementation details}
\label{sec:impl}

All arms use the same frozen checkpoint, prompt, parser, and deterministic
decoding, so a difference between arms cannot be attributed to a prompt or
parsing change; no weights are updated and no adapter is merged.  Missing
dimensions, malformed boxes, and invalid first-pass outputs are never
converted into a synthetic small box: they remain baseline misses and pass
through the router, crop, and fusion stages unchanged.  This preserves matched
denominators.  Appendix~\ref{app:repro} lists the exact prompt, parser
grammar, preprocessing constants, and the environment fields that were not
recoverable.

\subsection{Baselines and controls}

We compare the frozen default-resolution answer, the annotation-stratified
resolution diagnostic, conservative crop fusion without the label-free
router, and the complete composition.  The annotation-stratified arm is a
diagnostic rather than a deployable baseline: it uses ground-truth target
size only to quantify resolution headroom and is never a deployment
candidate, even though it reaches 89.22\%.  The crop/fusion-only arm tests
whether the router contributes beyond conservative crop fusion.  Ten further
closed controls -- including direct supervised adaptation, a released
third-party transfer checkpoint, a full-parameter fine-tune, iterative
unguarded zooming, and placement and self-consistency controls matched to
the crop arm's call count -- are reported in Appendix~\ref{app:controls};
the placement and self-consistency controls rule out that the crop's gain
comes merely from a second look rather than an informed crop location.
Every comparison is paired on identical expressions and resampled by image
cluster (Appendix~\ref{app:metrics}).  We present $\macc_{0.5:0.95}$ as
the manuscript's primary summary and Acc@0.5/0.75/0.9 alongside it.
This hierarchy is a retrospective presentation choice for Ref-L4; for frozen
transfer analyses, each table states its own Holm correction family.

\subsection{Statistical analysis}
\label{sec:statistics}

For every arm we compute paired row-level differences while resampling whole
images with replacement.  For an image $j$ carrying expressions
$i\in\mathcal{I}_j$, let $m_j^*$ be its multiplicity in a bootstrap draw.
The treatment effect is

\begin{equation}
  \widehat{\Delta}_t^{*}=
  \frac{\sum_j m_j^*\sum_{i\in\mathcal{I}_j}
  [u_t(\widehat b_i,y_i)-u_t(b_{0i},y_i)]}
  {\sum_j m_j^*|\mathcal{I}_j|},
  \label{eq:bootstrap}
\end{equation}

which estimates the mean effect for a randomly chosen expression while using
image clusters as the resampling units.  We use 10,000 bootstrap replicates
of \eqref{eq:bootstrap} and report two-sided percentile 95\% intervals on
$100\widehat{\Delta}_t$.  For the guard$\times$fusion factorial, the
per-dataset RefCOCO-family decomposition, and the IoU-threshold curve, every
metric within one comparison is computed from the same seeded resample at
each iteration, so cross-metric covariance is preserved; the remaining
per-arm interval tables use an independent, deterministic, metric-specific
seed per table.  The intervals reflect image clustering but not upstream
selection over candidate pipelines.  Secondary endpoints are an exploratory
family unless a table explicitly reports Holm adjustment.  Runtime and
invalid outputs are counted rather than dropped, and available row-level
reports retain paired win/loss/tie counts.  Appendix~\ref{app:metrics} gives
the complete metric definitions.

\section{Results}
\label{sec:results}

Table~\ref{tab:evidence-tier-summary} collects the deployed policy's effect
($CRG-B$) across the three evidence tiers before the per-evaluation detail
below.

\begin{table*}[t]
\centering
\small
\resizebox{\linewidth}{!}{%
\begin{tabular}{p{0.19\linewidth}p{0.11\linewidth}rrrr}
\toprule
Evidence tier & $n$ (expr / img) & $B$ Acc@0.5 & $CRG$ Acc@0.5 & $\Delta$Acc@0.5 [95\% CI] & $\Delta$mAcc [95\% CI] \\
\midrule
Ref-L4 (retrospective) & 31{,}921 / 9{,}467 & 88.531 & 89.725 & +1.194 [0.944, 1.439] & +3.066 [2.818, 3.311] \\
RefCOCO family (frozen transfer) & 30{,}969 / 3{,}982 & 89.212 & 90.029 & +0.817 [0.639, 1.001] & +0.645 [0.469, 0.827] \\
Flickr30K (prospective, merged-box) & 14{,}481 / 999 & 74.864 & 75.678 & +0.815 [0.299, 1.324] & +0.973 [0.648, 1.310] \\
\bottomrule
\end{tabular}%
}
\caption{Evidence-tier summary: $CRG$ against the frozen pre-resolution
incumbent $B$, same rows, image-cluster bootstrap intervals. Ref-L4 is
retrospective; RefCOCO family and Flickr30K are frozen/prospective transfer
that retuned nothing. Full metric detail is in
Table~\ref{tab:main-e1-per-dataset} and Appendix~\ref{app:evidence-tiers}.}
\label{tab:evidence-tier-summary}
\end{table*}

\subsection{Retrospective full-split matched comparison}

Table~\ref{tab:evidence-tier-summary} above and Table~\ref{tab:main-results}
in the appendix report this population's full metric set and one paired
denominator. The annotation-routed resolution diagnostic, reported
separately because it uses target size, reaches 89.220\% (+0.689 points);
it is neither deployable nor an upper bound, and the full deployable system
now exceeds it, since crop re-grounding and fusion contribute accuracy the
routing-only diagnostic cannot.

The larger differences at stricter thresholds are consistent with improved
boundary placement, but do not by themselves distinguish tighter boxes from
changes in target selection, annotation convention, or domain mix.  The
archived report records one invalid output per arm, 6,641 routed rows
(20.8\%), and 29,523 accepted-and-fused crop candidates of 31,920 attempts.
Appendix~\ref{app:m2-factorial} decomposes this into a guard $\times$
fusion factorial: the guard, not the midpoint rule, is what protects
Acc@0.5.

\subsection{Cross-dataset transfer to the RefCOCO family}
\label{sec:transfer}

After the primary Ref-L4 analysis was frozen, we applied the same
checkpoint, prompt, parser, router, resolution floor, crop guards, and
midpoint fusion to the five official RefCOCO/RefCOCO+/RefCOCOg test partitions
(30,969 expressions over 3,982 images), retuning nothing.  Every threshold and
weight was fixed before any RefCOCO-family row was scored, so this comparison
is prospective with respect to the method, unlike the retrospective Ref-L4
estimate of Section~\ref{sec:results}.

The pooled result (Table~\ref{tab:evidence-tier-summary}) is positive on
the primary endpoint and the operating threshold, flat at the strictest
one: Acc@0.75 gains $+0.853$pp ([0.566, 1.147]) and mean IoU gains
$+0.552$pp ([0.438, 0.668], $p<0.001$), each with a Holm-adjusted
$p\mathrel{\approx}0.003$ where reported.  Acc@0.9 does not move: $-0.029$
points ([$-0.554$, $0.497$], adjusted $p=0.93$).  The paired report records
13,925 IoU wins, 12,357 losses, and 4,687 ties, with two invalid rows in
each arm.  Table~\ref{tab:main-e1-per-dataset} breaks the deployed policy's
own effect, $CRG-B$, down by dataset: every dataset individually improves
on Acc@0.5, mAcc, and mean IoU, ten of the twelve dataset-by-metric
comparisons clear Holm-adjusted significance, and Acc@0.9 is
indistinguishable from zero on every dataset, matching the pooled null
rather than concealing a per-dataset regression.  A separate stage
decomposition (Appendix~\ref{app:e3-transfer}) explains why the pooled
Acc@0.9 null is not simply ``no effect'': routing alone gains $+1.162$
points at Acc@0.9, while the marginal effect of crop, guard, and fusion
after routing is $-1.192$ points relative to the routed arm, so the two
stages substantially offset.  Appendix~\ref{app:e3-transfer} also gives the
matching failure taxonomy.

\begin{table*}[t]
\centering
\small
\resizebox{\linewidth}{!}{%
\begin{tabular}{lrrrrrr}
\toprule
Dataset & Rows & Images & Acc@0.5 $\Delta$ & Acc@0.75 $\Delta$ & Acc@0.9 $\Delta$ & mAcc $\Delta$ \\
\midrule
RefCOCO & 10{,}752 & 1{,}500 & \textbf{+0.772} [0.516, 1.037] & +0.456 [0.037, 0.881] & $-0.530$ [$-1.301$, 0.234] & \textbf{+0.343} [0.083, 0.594] \\
RefCOCO+ & 10{,}615 & 1{,}500 & \textbf{+0.641} [0.381, 0.908] & \textbf{+0.895} [0.476, 1.319] & +0.377 [$-0.360$, 1.110] & \textbf{+0.644} [0.385, 0.911] \\
RefCOCOg & 9{,}602 & 2{,}600 & \textbf{+1.062} [0.751, 1.391] & \textbf{+1.250} [0.804, 1.705] & +0.083 [$-0.587$, 0.773] & \textbf{+0.984} [0.698, 1.281] \\
\bottomrule
\end{tabular}%
}
\caption{Per-dataset deployed-policy effects, $CRG-B$ (final policy against
the original pre-routing incumbent $B$), image-clustered 10,000-iteration
bootstrap, Holm-corrected within the 12-comparison (3 datasets $\times$ 4
metrics) family; bold marks adjusted $p<0.05$.  Row-weighted deltas equal
the pooled $CRG-B=(+0.817,+0.852,-0.029,+0.645)$ reported above, verified
by a machine-checked weighted-identity assertion (Appendix~\ref{app:e3-transfer}).
Image counts overlap across RefCOCO/RefCOCO+; RefCOCOg draws on a largely
disjoint image set.}
\label{tab:main-e1-per-dataset}
\end{table*}

\paragraph{The resolution router acts on small targets, and only on them.}
The label-free router flagged 2,223 of the 30,969 rows (7.18\%) for a
higher-resolution second pass, and changed the predicted box on 2,213 of them
(99.6\%).  Routing alone is significant on every metric: $+0.691$ points at
Acc@0.5 (95\% CI [0.554, 0.836]), $+0.807$ at Acc@0.75, $+1.162$ at Acc@0.9,
and $+0.861$ at mAcc, all with Holm-adjusted $p\mathrel{\approx}0.003$, from
1,444 IoU wins against 651 losses on the 2,095 rows whose accuracy
classification flips at some threshold (28,874 of the 30,969 rows tie; the
remaining 118 of the 2,213 coordinate changes move the box without crossing
any scoring threshold); Appendix~\ref{app:erratum} documents an earlier
resolution-router defect and its correction.

The size decomposition shows the effect is concentrated exactly where the
router is designed to act.  Against the incumbent, routing alone moves Acc@0.9
by $+10.13$ points on the small bucket (1,647 rows, 41.96\% to 52.09\%), by
$+0.93$ on the medium bucket, and negligibly on the large bucket (9,567
rows).  A floor that forces upscaling therefore buys strict-IoU accuracy in
proportion to how few pixels the target originally occupied, the mechanism
the router assumes rather than merely a correlate of it.

\paragraph{The guard is load-bearing, not a trade against a competitive
alternative.}
Because the crop candidate is retained before the guards and fusion are
applied, the stages separate on identical rows
(Table~\ref{tab:app-e3-arms}).  Crop re-grounding alone, with the guard still
active, is a boundary-precision mechanism: it gains $+1.844$ points at
Acc@0.9 (95\% CI [1.349, 2.350]) for only $+0.236$ at Acc@0.5.  Removing the
guard entirely -- a genuine ablation, recomputed directly from $CRG$'s own
stored candidate and guard fields with the guard gate removed -- does not
compound that gain: it underperforms the incumbent on every metric,
$-2.926$ points at Acc@0.5, $-3.969$ at Acc@0.75, and $-5.748$ at Acc@0.9
(Table~\ref{tab:app-components} gives the full intervals).  An archived run
under the same ``$CR$'' label uses different upstream predictions and is
therefore not a guard ablation; we retain it only as an archival artifact
with full provenance in Table~\ref{tab:app-e3-arms}.

The guards and midpoint fusion cost a small, real amount of strict-IoU
accuracy relative to the incumbent ($-0.029$ points at Acc@0.9, adjusted
$p=0.93$, not distinguishable from zero) in exchange for the largest Acc@0.5
gain of any arm ($+0.817$).  That is the only real trade in this
decomposition: guarding the crop candidate protects rows the incumbent
already had approximately right, which is where Acc@0.5 is won, at a
strict-IoU cost too small to distinguish from zero.  The paired per-row change confirms $CRG$ helps more rows than it harms
(13,291 versus 12,629 of 30,967 scoreable rows), though the harmed rows lose
slightly more IoU on average than the helped rows gain (mean IoU delta
$-0.001$) -- many small, mostly favorable boundary shifts plus a smaller
number of larger losses, not one mechanism dominating the other
(Appendices~\ref{app:e3-transfer} and~\ref{app:e5-qualitative}).

We ship one deployment policy, named explicitly rather than left implicit:
\textbf{LFPR} (guards plus midpoint fusion, arm $CRG$), the arm reported in
the abstract's headline numbers.  Removing the guard is not a viable
second policy: it costs no less than $CRG$ -- the guard is a zero-cost
accept/reject decision on an already-computed crop candidate, so
$CR^\dagger$ and $CRG$ issue identical model calls
(Table~\ref{tab:app-cost-r3}) -- and it is not competitive on any accuracy
metric (Table~\ref{tab:app-e3-arms}).

The guard$\times$fusion factorial (Appendix~\ref{app:m2-factorial}) shows
guarded replacement, an arm we did not ship, significantly beats $CRG$ at
Acc@0.9 and at $\macc_{0.5:0.95}$ -- our primary endpoint -- at the cost
of Acc@0.5; we retain midpoint fusion as the policy fixed before this
factorial was computed, and report the direct paired contrast in
Appendix~\ref{app:m2-factorial}.

\subsection{Comparison against released specialists: accuracy is not ordered,
it is redistributed}
\label{sec:specialists}

\begin{figure}[t]
\centering
\includegraphics[width=0.72\linewidth]{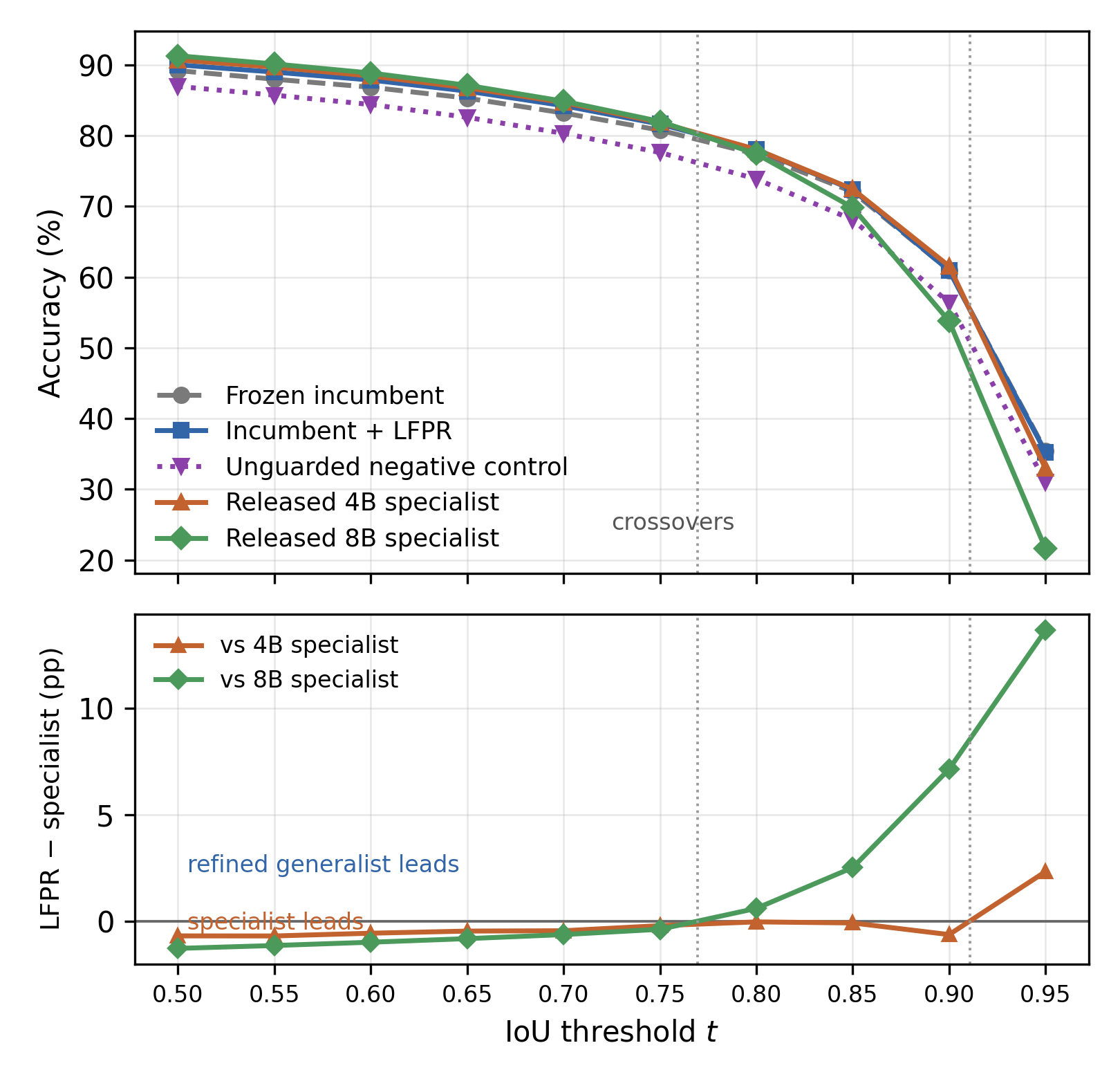}
\caption{Accuracy against IoU threshold on the 30,969-row transfer split
(top) and the paired difference between the guarded system and each
specialist (bottom).  Specialists lead at permissive thresholds; LFPR
crosses the 8B specialist near $t\approx0.78$ and widens thereafter, but
tracks the incumbent and the 4B specialist at strict thresholds until
$t=0.95$.  The unguarded negative control trails every system across the
full range, reaching only 30.99\% at $t=0.95$ against 35.25\% for LFPR,
35.43\% for the incumbent, and 32.93\%/21.60\% for the specialists.}
\label{fig:crossover}
\end{figure}

We also
evaluated two released referring-expression specialists, at 4B and 8B
parameters, zero-shot on exactly the same rows, clusters, and official
evaluator \citep{zhan2026egm}; both emit a single answer in one pass with no
refinement.  Because the specialists are supervised on this benchmark family
while our incumbent is zero-shot, we do not read this as a like-for-like
leaderboard contest; its purpose is to test whether the refinement remains
useful once the incumbent is already specialised.
Table~\ref{tab:specialists} reports the result and
Figure~\ref{fig:crossover} the full threshold profile.

\begin{table*}[t]
\centering
\small
\resizebox{\linewidth}{!}{%
\begin{tabular}{@{}p{0.30\textwidth}rrrrrrr@{}}
\toprule
System & Params. & Acc@0.5 & Acc@0.75 & Acc@0.9 & mAcc & Mean IoU & Invalid \\
\midrule
Frozen incumbent & 8B & 89.212 & 80.790 & 60.955 & 75.925 & 82.262 & 2 \\
\quad + resolution routing & 8B & 89.903 & 81.597 & \textbf{62.117} & \textbf{76.786} & 82.911 & 2 \\
\quad + LFPR & 8B & 90.029 & 81.643 & 60.926 & 76.570 & 82.814 & 2 \\
Released specialist & 4B & 90.723 & 81.856 & 61.558 & 76.722 & 83.171 & 115 \\
\quad + LFPR & 4B & 91.007 & 82.311 & \textbf{63.128} & \textbf{77.508} & \textbf{83.592} & 115 \\
Released specialist & 8B & 91.311 & 82.027 & 53.786 & 74.705 & 82.365 & 0 \\
\quad + LFPR & 8B & \textbf{91.688} & \textbf{83.257} & 60.502 & 77.106 & 83.583 & 0 \\
\midrule
\quad + crop, routing, \emph{no guard} ($CR^\dagger$, negative control) & 8B &
86.974 & 77.632 & 56.382 & 72.707 & 79.974 & 2 \\
\bottomrule
\end{tabular}%
}
\caption{Zero-shot released specialists against the frozen incumbent and
LFPR, each also shown with LFPR applied to its own box (every delta
significant; CIs in Appendix~\ref{app:egm-lfpr-accuracy}).  LFPR improves
every specialist on every metric including Mean IoU, most sharply the 8B
specialist's Acc@0.9
($+6.72$).  Among the base systems, best-Acc@0.5 (8B specialist) has worst
Acc@0.9/mAcc; the trade persists after refinement.  The bottom row is a
negative control: the guard gate removed from $CRG$'s own stored
candidates.  It underperforms every system above it on every metric.
Invalid predictions remain in every denominator.}
\label{tab:specialists}
\end{table*}

The headline reading is unfavourable to our method: a released 4B model doing
one forward pass exceeds the refined 8B pipeline by $0.694$ points at Acc@0.5
(95\% CI $[0.234, 1.148]$, adjusted $p=0.036$), and the 8B specialist exceeds
it by $1.282$ points ($[0.883, 1.698]$).  Refining a frozen generalist
narrows but does not close the gap to a task-specialised model at the
permissive threshold that dominates published comparisons.  The
strict-threshold reading reverses, but only for the 8B specialist: the
guarded system is indistinguishable from the 4B specialist at Acc@0.9
($-0.633$ points, 95\% CI $[-1.545, 0.228]$, adjusted $p=0.96$) and beats the
8B specialist by $+7.139$ points ($[6.169, 8.125]$) with $+1.865$ points of
mAcc.  A released 4B specialist matches the guarded system at every
threshold we report and beats it at Acc@0.5; only against the 8B specialist
does the guarded system's strict-IoU advantage show up.

The same negative control (Table~\ref{tab:specialists}, bottom row) trails
both specialists at Acc@0.5 by a wider margin than the guarded system and no
longer beats either convincingly at Acc@0.9, confirming the 8B-specialist
reversal is a property of the guarded system specifically.  Scaling the
specialist from 4B to 8B sharpens the same trade-off: Acc@0.5 rises
$+0.588$ points while Acc@0.9 falls $7.772$ and mAcc falls $2.017$
(Appendix~\ref{app:relocated} gives the full row-level detail).  An
image-level firewall audit finds zero training-image overlap between the
specialists and any evaluation partition, so their Acc@0.5 advantage is
genuine rather than memorization, though benchmark-specific: the direction
reverses on long expressions (Appendix~\ref{app:relocated}).

\paragraph{Refinement composes with specialisation.}
Running the complete frozen procedure, unchanged, on each specialist's own
predictions over the same 30,969 rows improves every reported endpoint for
both (full intervals in Table~\ref{tab:egm-lfpr-accuracy}), with the
strict-threshold gain far larger for the stronger specialist, so the
benefit scales with the underlying model.

\subsection{A prospective Flickr30K Entities confirmation under the standard
merged-box protocol}
\label{sec:flickr30k}

A stronger test of generalisation is a dataset the method never saw during
development.  We use the official Flickr30K Entities test split
\citep{plummer2015flickr30k} under its standard merged-box protocol: every
visible phrase with at least one annotated box is represented by the enclosing
box of all its annotations.  The resulting 14,481-row, 999-image denominator
(Table~\ref{tab:evidence-tier-summary}; Appendix~\ref{app:flickr-merged}
gives the manifest and firewall detail) improves every endpoint:
Acc@0.75 rises $+1.326$ ($p=0.0008$), and Acc@0.9 rises $+1.022$
($p=0.0052$), with zero invalid predictions.

The stage decomposition is less uniformly favorable than the endpoint
comparison.  The baseline (B) incumbent reaches mAcc 56.975 and Acc@0.9
37.753; routing alone ($R$) reaches mAcc 58.585 and Acc@0.9 39.749; full
LFPR ($CRG$) reaches mAcc 57.948 and Acc@0.9 38.775.  Thus routing
transfers prospectively, but crop/guard/fusion does not add accuracy beyond
routing on this dataset; its marginal value is dataset-dependent.  The guard
nevertheless remains load-bearing because the matched unguarded arm falls
below the original incumbent (Appendix~\ref{app:flickr-mechanism}).

As a sensitivity analysis on the same images, we also retain the historical
single-box selection: excluding 2,907 multi-box and 3,038 non-visual phrases
leaves 11,574 rows over 979 images.  It yields larger changes (mAcc
$+2.575$, Acc@0.5 $+1.175$, Acc@0.75 $+3.145$, Acc@0.9 $+3.689$), but is not
an independent confirmation.  Both protocols were frozen before scoring,
and an image-hash audit finds no overlap with historical development data.

\section{Analysis and Discussion}
\label{sec:discussion}

\subsection{Why the stages can oppose one another, and what a single threshold hides}
\label{sec:sota}

The resolution stage changes the visual evidence available to the model;
the crop stage changes the coordinate frame in which it localizes; their
effects are separable and sometimes opposing
(Table~\ref{tab:main-e1-per-dataset}; Table~\ref{tab:app-e3-arms}).
Resolution routing alone earns a clean, positive strict-IoU effect on
every tier, accounting for most of the deployed policy's Acc@0.9 effect
(Appendix~\ref{app:refl4-decomposition}).  Crop re-grounding, the guard,
and fusion protect against an unfiltered second opinion rather than
guarantee further accuracy: on the RefCOCO family and Flickr30K
(Appendix~\ref{app:flickr-mechanism}) their marginal effect beyond
routing trades Acc@0.5 for Acc@0.9/mAcc, never a net gain; Ref-L4 is the
exception, but it is also the most heavily tuned tier, so only the
untuned tiers speak without a tuning confound.  The robust claim is
narrower than ``components compose'': selective resolution transfers,
the policy improves broadly, and the guard prevents an unfiltered crop
answer from becoming competitive.

The specialist comparison makes a methodological point that does not
depend on our method: reporting Acc@0.5 alone can invert the ordering of
systems relative to their localization quality, since two systems can be
separated by seven points at Acc@0.9 while differing by one point the
other way at Acc@0.5, and only the latter is usually reported.  We
therefore recommend reporting referring-expression results as a threshold
profile, or at minimum with mean accuracy alongside Acc@0.5.

\subsection{Limitations}

Our evidence has five boundaries.

First, the Ref-L4 estimate is retrospective: the published test
distribution shaped diagnostics and selection, pilot rows were reused, and
no external preregistration exists, so its intervals describe sampling
variability under a frozen comparison, not selection across the
pipelines, prompts, resolutions, guards, and fusion variants explored.
The RefCOCO-family result is prospective, with its own scoped Holm
correction.

Second, the specialist comparison is only partly compute-normalised
(Appendix~\ref{app:relocated} gives same-host latency ratios); energy and
FLOPs were not measured.

Third, the specialists are supervised on the RefCOCO family while our
incumbent is zero-shot; a firewall audit finds no training-image overlap,
so the comparison is uncontaminated but not like-for-like.

Fourth, our operator-effect evidence covers one architecture family:
LFPR improves every reported endpoint on both released specialists and a
$2\times$-smaller checkpoint (Appendix~\ref{app:p0-design}), but
cross-family checkpoints were screened only at the direct-answer stage
(Appendix~\ref{app:cross-family}).

Fifth, the size threshold, guards, and midpoint weight are fixed
constants: a local grid sweep finds no configuration that Pareto-dominates
the shipped rule (Appendix~\ref{app:controls}), and we did not learn them,
since that would change the contribution from a frozen operator to a
learned router.

\section{Conclusion}
\label{sec:conclusion}

We studied a narrow deployment question: can a frozen grounding-capable VLM
use its own incumbent prediction to obtain a more precise box without
accessing target annotations at inference?  LFPR's lesson is not that a
second prediction is uniformly better: components move different,
opposing parts of the curve, and an ablation confirms the guard is
load-bearing: a complement to stronger grounding models, not a
replacement.

\section*{Reproducibility Statement}

The complete source code implementing LFPR and the analysis scripts used
to produce every table and figure in this paper are publicly available at
\url{https://github.com/mabo1215/BodhiNet}.  All prompts, pixel budgets,
clipping conventions, guard thresholds, and decoding settings used by
LFPR are given in Appendix~\ref{app:repro}, together with the
preprocessing constants and the environment fields that could not be
recovered from archived logs.  The image-cluster bootstrap procedure and
Holm correction used for every reported interval are specified in
Appendix~\ref{app:metrics}.  The complete label-free procedure is stated
as pseudocode in Appendix~\ref{app:algorithm}.  Dataset splits, source
labels, and the data-firewall audits establishing image-disjointness
between evaluation partitions and prior development data are described in
Section~\ref{sec:protocol} and Appendix~\ref{app:data-governance}, which
also gives the SHA-256 checksum of the exact per-row prediction file
behind every headline result (Table~\ref{tab:app-artifact-manifest}); the
released code includes the full-length hashes and generation commands in
a machine-readable manifest.

\bibliographystyle{plainnat}
\bibliography{references}

\clearpage
\appendix

This supplement gives the recoverable implementation specification, data-use
audit, detailed result intervals, and disposition of the matched control
experiments for LFPR.  Missing provenance fields are stated explicitly rather
than reconstructed from unversioned memory.

\section{Reproducibility Specification}
\label{app:repro}

The complete source code implementing the LFPR pipeline -- the resolution
router, crop re-grounding, geometric guards, midpoint fusion, and the
analysis scripts used to produce every table and figure in this paper -- is
publicly available at \url{https://github.com/mabo1215/BodhiNet}.

\subsection{Model and decoding}

The generalist arms use a local copy of \texttt{Qwen/Qwen3-VL-8B-Instruct}
at a single pinned revision, verified identical across every downloaded
file's cache metadata (the exact revision hash is recorded in the released
run manifest rather than restated here), in bfloat16, SDPA attention,
evaluation mode, and deterministic, non-sampling greedy decoding, at most
64 generated tokens for the full-split accuracy runs.  The synchronized
cost protocol raises the ceiling to 4,096
new tokens and reports the realized generated-token count.  The full-split
Ref-L4, RefCOCO-family, and Flickr30K accuracy runs, and the policy-matched
cost audit, used Python 3.10.12, PyTorch 2.8.0 (CUDA 12.8 build),
Transformers 4.57.1, Pillow 11.1.0, NumPy 1.26.4, NVIDIA driver 595.84,
Ubuntu 22.04.5 LTS, kernel 5.15.0, on a four-GPU host (NVIDIA GeForce RTX
3090, 24~GiB each).  The full-parameter fine-tune control ran separately
on an NVIDIA RTX PRO 6000 Blackwell Server Edition with approximately
96~GiB of memory; earlier development-phase pilots not reported as final
numbers in this paper used a mix of rented cloud GPU instances, documented
per-experiment in the released run logs rather than restated here.

The specialist cost audit uses local EGM-4B and EGM-8B checkpoints in
bfloat16 with SDPA attention and greedy decoding, allowing at most 4,096 new
tokens.  It ran on a host with four physical NVIDIA RTX~3090 GPUs (24~GiB
each).  Each baseline/refinement arm was divided into two non-overlapping row
shards, with one process per GPU; no GPU hosted more than one worker.

\subsection{Determinism check}
\label{app:determinism}

Since LFPR uses frozen weights and greedy decoding, training-seed
replication is not meaningful; instead we verify output-level determinism
directly. A fixed 128-row deterministic slice of the direct-answer
incumbent (arm $B$) was run in three independent processes launched in
parallel on three separate physical GPUs on the same host (rather than
three sequential repeats on one card, which additionally rules out
cross-device, not just cross-process, non-determinism). All three runs'
predicted boxes were byte-identical on $128/128$ rows: a $100\%$
reproducibility rate under greedy decoding on this hardware.

The full-image prompt is, verbatim:
\begin{quote}\small
\textit{Locate the region that matches: <expression>. Reply only with its
bounding box as [x1,y1,x2,y2] in coordinates normalized to 0--1000. No
explanation.}
\end{quote}
The crop-pass prompt replaces the final two sentences:
\begin{quote}\small
\textit{Reply only with its bounding box as [x1,y1,x2,y2] in coordinates
normalized to 0--1000 within this crop.}
\end{quote}
In both cases the checkpoint's native chat template is applied with an
image followed by the user text, with the chat template's generation-prompt
formatting enabled; no system message or stop string is added.

The default first-pass image-processing configuration is loaded from the
checkpoint.  The routed pass overrides the minimum-pixel budget to
4,194,304 and leaves the checkpoint's maximum unchanged.  The crop pass
reuses the same 4,194,304-pixel minimum override and also leaves the
checkpoint maximum unchanged, which resolves to an effective processed-image
size bounded between 4,194,304 and 16,777,216 pixels for both passes; this
was confirmed by reading the image processor's own logged effective
configuration from the run that produced every crop-based result in this
paper, not from the smaller, crop-specific pixel range (approximately
200,704 to 1,003,520 pixels) that the runner accepts as an argument but
never actually applies.  Source images
are opened with PIL, honor PIL's decoded orientation as stored, and are
converted to RGB; explicit EXIF transposition was not applied.  Crops use PIL's
default resampling path through the Qwen processor.  No weights are updated or
adapter merged.

The parser strips the completion, accepts exactly one JSON/bare four-number
box (a singleton wrapper is permitted), and rejects zero or multiple boxes.
Coordinates must satisfy $0\le x_1<x_2\le1000$ and
$0\le y_1<y_2\le1000$ before division by 1000.  Invalid, non-finite,
zero-area, or inverted boxes score as misses; no row is dropped.  Crop pixel
bounds use floor/ceiling, are clipped to the source image, and the accepted
candidate is mapped back without integer rounding.

\subsection{Router and crop specification}

The normalized incumbent box is converted to original-image pixel
coordinates.  If $s=\sqrt{wh}<128$ pixels, the row is routed to the
high-resolution pass; $128\leq s\leq256$ is medium and is not routed; all
larger boxes are not routed.  Missing dimensions, malformed boxes, and
invalid first-pass outputs are never converted into a synthetic small box.
They remain baseline misses and are recorded in the error counters.

For every valid post-resolution incumbent, the crop width and height are each
twice the incumbent width and height, capped at one and shifted to remain
inside the image.  Zoom consistency is the Euclidean distance between the crop
candidate center and the incumbent center expressed in crop coordinates,
divided by $\sqrt{1/2}$.  The candidate is admitted only when this value is at
most 0.35, candidate--incumbent IoU is at least 0.25, and the area ratio lies
in $[0.5,2.0]$.  The output is the coordinate-wise midpoint.  Invalid
incumbents short-circuit as misses; invalid high-resolution or crop answers,
zero-area denominators, degenerate crops, and runtime errors never trigger
fusion.  Available history says these constants preceded the full-split run,
but not that they preceded all access to the published test distribution.

\subsection{Data firewall}
\label{app:data-governance}

The original ``test firewall'' claim is withdrawn.  The 2,000-row pilot is a
deterministic image-disjoint holdout relative to named earlier pilots, but it
was sampled from the already-scored Ref-L4 test prediction file and reused by
the crop/fusion-only and full LFPR arms.  The complete published test split was also used for annotation-size
and source diagnostics that influenced later choices.  Thus the holdout is a
useful same-row engineering check, not an untouched confirmation benchmark.

\begin{table*}[t]
\centering
\footnotesize
\setlength{\tabcolsep}{2pt}
\begin{tabularx}{\linewidth}{@{}p{0.18\linewidth}rrp{0.12\linewidth}p{0.16\linewidth}X@{}}
\toprule
Artifact & Rows & Images & Label access & First recoverable access & Recorded uses and decisions \\
\midrule
Ref-L4 published test & 31,921 & 9,467 & Yes & 2026-08-05 or earlier & Baseline scoring; source/size diagnostics; resolution arm; final LFPR scoring and manuscript framing. \\
Frozen RefCOCO-family transfer & 30,969 & 3,982 & Yes & 2026-08-10 & Frozen LFPR pooled audit; descriptive dataset slices; no target-label tuning. \\
Shared crop/refinement holdout & 2,000 & 1,713 & Labels present in rows & 2026-08-08 & Selected from the full-test baseline after excluding named prior-pilot images; reused for crop/fusion-only and full LFPR decisions. \\
Resolution/prompt pilots & 800 each & Not fully exported & Yes & 2026-08-07--08 & Floor, prompt, and stopping choices; some samples were reused and the rung summaries used different references. \\
Supervised-adaptation split & 3,314 & Not fully exported & Yes & 2026-08-08 & Supervised-adaptation control and stopping decision; relation to all other image pools is not reconstructable from the compact export. \\
Backbone-scale pilot & 380 & 344 & Yes & 2026-08-09 & Same-family 32B direct-model control; no complete LFPR run. \\
\bottomrule
\end{tabularx}
\caption{Minimum data-governance record recoverable from compact manifests and
logs.  ``Not fully exported'' is a provenance gap, not evidence of separation.}
\label{tab:data-governance}
\end{table*}

Reproducible future evaluations require stable record keys, content hashes
rather than filenames, artifact checksums, and an access ledger with actor, purpose, split,
label availability, and decision influence.  Until that ledger and an
untouched benchmark/model combination are available, all reported intervals
are retrospective.

\paragraph{Artifact hash manifest.} Table~\ref{tab:app-artifact-manifest}
lists the checksum of the exact per-row prediction file behind each of this
paper's primary confirmatory results (Ref-L4, the RefCOCO family, and the
Flickr30K merged-box confirmation, including its 512-row matched subset).
Every downstream table that reports one of these arms' numbers is
traceable to exactly one of these hashes -- in particular, the RefCOCO
family's $B$ row is the artifact that Blocker~1's investigation confirmed
is genuinely pre-resolution, not the differently-named file that a prior
draft mistook for it. Hashes are truncated to 12 hex characters here for
space; the full 64-character digests, the exact generation commands, and
the artifacts still outside this manifest's coverage (the RefCOCO-family
cost audit and several secondary diagnostic tables) are listed in the
code supplement's machine-readable manifest referenced in the
Reproducibility Statement.

\begin{table}[t]
\centering
\small
\begin{tabular}{llrl}
\toprule
Dataset & Arm & Rows & SHA-256 (12 hex) \\
\midrule
Ref-L4 & $B$ & 31{,}921 & \texttt{a31497f9f3cb} \\
Ref-L4 & $R$ (diagnostic) & 31{,}921 & \texttt{c4d83a2c7800} \\
Ref-L4 & $CRG$ & 31{,}921 & \texttt{3bee48fa7c3f} \\
RefCOCO family & $B$ & 30{,}969 & \texttt{0b0e6a9b508d} \\
RefCOCO family & $CRG$ & 30{,}969 & \texttt{fe4508dff83b} \\
Flickr merged-box & $B$ & 14{,}481 & \texttt{24170b518493} \\
Flickr merged-box & $R$ & 14{,}481 & \texttt{1044ef57c8b0} \\
Flickr merged-box & crop-on-$R$ (raw) & 14{,}481 & \texttt{ed88edbf7f8c} \\
Flickr merged-box & $CRG$ & 14{,}481 & \texttt{f3be68709a53} \\
Flickr merged-box, 512-row matched & $B$ & 512 & \texttt{cc29a5e2f446} \\
Flickr merged-box, 512-row matched & crop-on-$R$ (raw) & 512 & \texttt{129cd8e0cb17} \\
Flickr merged-box, 512-row matched & $R_{\mathrm{ALL}}$ & 512 & \texttt{8ccaa327e65f} \\
Flickr merged-box, 512-row matched & $CR_{\mathrm{ALL}}$ & 512 & \texttt{e536db5e0514} \\
Flickr merged-box, full population & $R_{\mathrm{ALL}}$ & 14{,}481 & \texttt{41df2c56899c} \\
Flickr merged-box, full population & $CR_{\mathrm{ALL}}$ & 14{,}481 & \texttt{1db4562f197e} \\
\bottomrule
\end{tabular}
\caption{Headline artifact hash manifest. Each row is one frozen per-row
prediction file (record\_key, gt\_box, pred\_box, and arm-specific fields);
every headline table or figure that reports the corresponding arm's
accuracy or cost derives from exactly this file, not a superseded or
differently-scored copy of it.}
\label{tab:app-artifact-manifest}
\label{app:artifact-manifest}
\end{table}

\section{Detailed Results}
\label{app:detailed-results}

\paragraph{Arm glossary.} Throughout this appendix, $B$ denotes the frozen
incumbent's pre-resolution prediction; $R$ denotes the prediction after
label-free resolution routing alone; $C$ denotes a guarded crop pass without
routing; $CR^\dagger$ denotes the unguarded negative control (routing plus
crop with the geometric guard removed); and $CRG$ denotes the full shipped
policy (routing, guarded crop, and midpoint fusion). A contrast written
$X{-}Y$ always means arm $X$ minus arm $Y$ on the same rows; tables and
prose state their reference arm explicitly because $CRG{-}B$ (the deployed
policy's effect) and $CRG{-}R$ (the post-routing stage effect) answer
different questions and are not interchangeable, as Table~\ref{tab:app-e1-per-dataset}
and its surrounding discussion make explicit.

\subsection{Evidence-tier summary}
\label{app:evidence-tiers}

Table~\ref{tab:evidence-tiers} collects the treatment-minus-incumbent effect
on every primary and secondary endpoint across the paper's three evidence
tiers, before the per-evaluation breakdowns below.  Effects are differences
against each evaluation's own matched incumbent; the Ref-L4 and RefCOCO-family
rows draw from Tables~\ref{tab:main-results} and~\ref{tab:app-e3-transfer},
and the Flickr rows from Section~\ref{sec:flickr30k} of the main text and
Appendix~\ref{app:flickr-merged}.  Mean-IoU deltas are in percentage points
and are given only where an archived report computed them; a blank cell
means the quantity was not computed, not that it was zero.

\begin{table*}[t]
\centering
\small
\resizebox{\linewidth}{!}{%
\begin{tabular}{p{0.20\linewidth}p{0.19\linewidth}rrrrrr}
\toprule
Evaluation & Evidence tier & $n$ (expr / img) & $\Delta$Acc@.5 & $\Delta$Acc@.75 & $\Delta$Acc@.9 & $\Delta$mAcc & $\Delta$Mean IoU \\
\midrule
Ref-L4 (full split) & Retrospective development & 31{,}921 / 9{,}467 & +1.194 & +3.173 & +5.354 & +3.066 & +1.704 \\
RefCOCO/+/g (pooled) & Frozen cross-dataset transfer & 30{,}969 / 3{,}982 & +0.817 & +0.852 & $-0.029$ & +0.645 & +0.552 \\
Flickr30K (merged-box) & Prospective, image-disjoint & 14{,}481 / 999 & +0.815 & +1.326 & +1.022 & +0.973 & +0.712 \\
Flickr30K (single-box) & Prospective, image-disjoint (secondary) & 11{,}574 / 979 & +1.175 & +3.145 & +3.689 & +2.575 & -- \\
\bottomrule
\end{tabular}%
}
\caption{Evidence-tier summary of the three headline evaluations, in
percentage points against each row's own matched incumbent.  Ref-L4 is
retrospective development evidence and should not be read as confirmatory;
the RefCOCO-family pooled row is frozen transfer that retuned nothing;
Table~\ref{tab:app-e1-per-dataset} decomposes this same $CRG-B$ effect by
dataset; the two Flickr30K rows are the paper's
prospective, image-disjoint confirmation, sharing the same 999/979 source
images under two different phrase-to-box protocols (Appendix~\ref{app:flickr-merged}) rather than constituting two
independent datasets.}
\label{tab:evidence-tiers}
\end{table*}

\subsection{Pilot and full-split intervals}

\begin{table*}[t]
\centering
\small
\begin{tabular}{lrrrrrr}
\toprule
Arm & $n$ & Acc@0.5 & Acc@0.75 & Acc@0.9 & mAcc & Objects365 $\Delta$ \\
\midrule
Crop/fusion reused pilot & 2,000 & +0.70 & -- & -- & -- & -- \\
LFPR reused pilot & 2,000 & +0.85 & +2.65 & +2.65 & +2.055 & +1.30 \\
LFPR full test & 31,921 & +1.194 & +3.173 & +5.354 & +3.066 & +1.720 \\
\bottomrule
\end{tabular}
\caption{Historical treatment-minus-baseline effects in percentage points.
The two pilot rows are reused and are not independent confirmation.  LFPR pilot
one-sided LCBs are +0.45, +1.85, +1.61, +1.69, and +0.68 for Acc@0.5,
Acc@0.75, Acc@0.9, mAcc, and Objects365 Acc@0.5, respectively.  The full
test Acc@0.5 paired image-level 95\% interval is [+0.944,+1.439].
LCBs explain historical selection and are not confirmatory.}
\label{tab:app-d9-intervals}
\end{table*}

\subsection{Cross-dataset transfer to the RefCOCO family}
\label{app:e3-transfer}

The transfer evaluation applies the frozen procedure without retuning to five
official RefCOCO/RefCOCO+/RefCOCOg test partitions.  The pooled paired audit
uses 30,969 expressions, 3,982 image clusters, exact record-key joins, and
10,000 image-cluster bootstrap resamples.  These partitions influenced no
design decision: every threshold, guard, resolution floor, and fusion weight
was fixed on Ref-L4 before any row here was scored.

\begin{table*}[t]
\centering
\small
\begin{tabular}{lrrrrr}
\toprule
Metric & Incumbent & LFPR & $\Delta$ (pp) & Two-sided 95\% CI & Holm $p$ \\
\midrule
Acc@0.5 & 89.212 & 90.029 & +0.817 & [0.639, 1.001] & 0.0032 \\
Acc@0.75 & 80.790 & 81.643 & +0.852 & [0.566, 1.147] & 0.0032 \\
Acc@0.9 & 60.955 & 60.925 & $-0.029$ & [$-0.554$, 0.497] & 0.9311 \\
mAcc@0.5:0.95 & 75.925 & 76.570 & +0.645 & [0.469, 0.827] & 0.0032 \\
Mean IoU & 82.262 & 82.814 & +0.552 & [0.438, 0.668] & --- \\
\bottomrule
\end{tabular}
\caption{Pooled transfer audit on the RefCOCO/RefCOCO+/RefCOCOg family, after
the resolution floor was verified to take effect.  Intervals are
image-cluster bootstrap intervals from the archived paired report; the Holm
value is the multiplicity-adjusted primary-family value.  Acc@0.9 is the one
metric that does not move: the guards and midpoint return the strict-IoU
accuracy that the crop and floor earn (Table~\ref{tab:app-e3-arms}).  Mean
IoU is outside the reported four-metric Holm family (no Holm-adjusted $p$
reported here, though its own bootstrap $p<0.001$) and moves in the
\emph{same} direction as Acc@0.5/mAcc, unlike the strict-IoU metric --
consistent with the deployed policy broadly improving boundary precision
even though its strict-threshold effect nets to zero after the crop stage
gives back part of routing's Acc@0.9 gain (Table~\ref{tab:app-e3-arms}).
Each arm contains two invalid rows, which remain in the denominator.}
\label{tab:app-e3-transfer}
\end{table*}

The paired report records 13,925 IoU wins, 12,357 losses, and 4,687 ties.
Table~\ref{tab:app-e1-per-dataset} decomposes this pooled full-policy
result, $CRG-B$ (the deployed policy against the original pre-routing
incumbent $B$), by dataset, using the same exact-key join and image-cluster
bootstrap as the pooled table, verified by a machine-checked assertion to
reproduce the pooled estimate under row weighting to within $10^{-10}$
(Appendix~\ref{app:metrics}).

\begin{table*}[t]
\centering
\small
\resizebox{\linewidth}{!}{%
\begin{tabular}{lrrrrrr}
\toprule
Dataset & Rows & Images & Acc@0.5 $\Delta$ & Acc@0.75 $\Delta$ & Acc@0.9 $\Delta$ & mAcc $\Delta$ \\
\midrule
RefCOCO & 10{,}752 & 1{,}500 & \textbf{+0.772} [0.516, 1.037] & +0.456 [0.037, 0.881] & $-0.530$ [$-1.301$, 0.234] & \textbf{+0.343} [0.083, 0.594] \\
RefCOCO+ & 10{,}615 & 1{,}500 & \textbf{+0.641} [0.381, 0.908] & \textbf{+0.895} [0.476, 1.319] & +0.377 [$-0.360$, 1.110] & \textbf{+0.644} [0.385, 0.911] \\
RefCOCOg & 9{,}602 & 2{,}600 & \textbf{+1.062} [0.751, 1.391] & \textbf{+1.250} [0.804, 1.705] & +0.083 [$-0.587$, 0.773] & \textbf{+0.984} [0.698, 1.281] \\
\bottomrule
\end{tabular}%
}
\caption{Per-dataset deployed-policy effects, $CRG-B$, image-clustered
10,000-iteration bootstrap, Holm-corrected within the 12-comparison
(3 datasets $\times$ 4 metrics) family. Bold marks Holm-adjusted $p<0.05$.
Row-weighted deltas equal the pooled $CRG-B=(+0.817,+0.852,-0.029,+0.645)$
in Table~\ref{tab:app-e3-transfer} to within $10^{-10}$.  Image counts
overlap across RefCOCO/RefCOCO+ (both built on the same COCO subset with
different expression styles); RefCOCOg draws on a largely disjoint image
set, which is why the three counts do not sum to the pooled 3,982.}
\label{tab:app-e1-per-dataset}
\end{table*}

The deployed policy improves every dataset individually at Acc@0.5, mAcc,
and mean IoU (not shown in the table; $+0.323$/$+0.518$/$+0.847$ points for
RefCOCO/RefCOCO+/RefCOCOg respectively, each $p<0.001$), and at Acc@0.75 for
RefCOCO+ and RefCOCOg; ten of the twelve dataset-by-metric comparisons in
the table clear Holm-adjusted significance.  Acc@0.9 is indistinguishable
from zero on every dataset (RefCOCO $-0.530$, RefCOCO+ $+0.377$, RefCOCOg
$+0.083$, none significant), matching the pooled null rather than
concealing a per-dataset regression -- no dataset shows a significant
$CRG-B$ regression on any reported metric.

A separate, secondary decomposition asks a different question: relative to
the \emph{post-routing} incumbent $R$ rather than $B$, does crop/guard/fusion
preserve the strict-IoU gain that routing alone earns?  Reusing the same
per-row artifact, $CRG-R$ (not shown as a full table) is $-1.423$
($[-2.206,-0.673]$, Holm-significant), $-0.537$ ($[-1.272,0.171]$, not
significant), and $-1.656$ ($[-2.288,-1.019]$, Holm-significant) at Acc@0.9
for RefCOCO/RefCOCO+/RefCOCOg, row-weighting to the pooled
$CRG-R=(+0.126,+0.046,-1.192,-0.216)$ for Acc@0.5/0.75/0.9/mAcc
(Table~\ref{tab:app-e3-arms}).  This explains the mechanism behind the
pooled $CRG-B$ Acc@0.9 null in Table~\ref{tab:app-e3-transfer} -- routing
gains strict-IoU accuracy that the later crop/guard/fusion stage partly
returns -- without changing the per-dataset $CRG-B$ conclusion above: $CRG-B$
and $CRG-R$ answer different questions (deployed-policy effect vs.\
post-routing marginal effect) and are not interchangeable.

\paragraph{What kind of harm this is.} We classified every harmed row on
the RefCOCO-family transfer split (final box lower IoU-vs-ground-truth than
the true pre-resolution incumbent $B$; $12{,}357$ of $30{,}969$ rows,
matching the paired loss count above) into one of six geometric categories
-- router-caused, crop-boundary truncation, referent switch (final box
barely overlaps its own pre-crop anchor), over-shrink, over-expand, or a
residual modest-perturbation category -- checked in that priority order per
row, using each row's already-recorded box geometry (no new inference).
The result is dominated by, but not exclusively, modest perturbation:
$94.58\%$ of harmed rows ($11{,}687/12{,}357$) are boundary shifts that
still overlap their pre-crop anchor substantially, $0.78\%$ are
over-expansions, and $0.11\%$ are over-shrinkages, with zero crop-boundary
truncations or referent switches.  A non-trivial remainder, $4.53\%$
($560/12{,}357$), is router-caused: the resolution router changed the
incumbent's box and that change alone, before the crop stage ever ran, was
already worse than the original pre-routing prediction.  The Acc@0.9 null
documented above is therefore mostly the aggregate of many small coordinate
perturbations from midpoint fusion crossing the strict threshold (the same
mechanism Table~\ref{tab:app-m2-factorial} isolates as the fusion rule's
own cost), plus a smaller contribution from the router itself occasionally
routing a row to a worse prediction -- not evidence of crop-boundary
truncation or referent switching.

The pooled read above could in principle hide a dataset-specific failure
mode that a pooled classification averages away. Table~\ref{tab:app-per-dataset-taxonomy}
repeats both the taxonomy and the strict-IoU transition-matrix cells on
each of RefCOCO/RefCOCO+/RefCOCOg individually rather than pooled. The
boundary-shift majority and near-zero crop-truncation/referent-switch rates
hold on every sub-dataset, but the router-caused share varies more than the
pooled figure suggests -- from $3.1\%$ on RefCOCO+ to $7.4\%$ on
RefCOCOg -- and strict-IoU harm rates cluster in a narrower
$5.7$--$6.5\%$ range. The pooled numbers are therefore a fair summary of
the dominant boundary-shift mechanism, but they average over real
dataset-level variation in how often the router itself is the source of
harm.

\begin{table}[t]
\centering
\small
\begin{tabular}{lrrrrr}
\toprule
Dataset & Boundary shift & Router-caused & Over-expand & Over-shrink & Strict harm rate$^\dagger$ \\
\midrule
RefCOCO & $95.76\%$ & $3.41\%$ & $0.72\%$ & $0.11\%$ & $6.48\%$ \\
RefCOCO+ & $95.96\%$ & $3.10\%$ & $0.80\%$ & $0.15\%$ & $5.70\%$ \\
RefCOCOg & $91.66\%$ & $7.44\%$ & $0.82\%$ & $0.08\%$ & $5.71\%$ \\
\bottomrule
\end{tabular}
\caption{Failure taxonomy and strict-IoU harm rate computed independently
on each RefCOCO-family sub-dataset (not pooled), against the true
pre-resolution incumbent $B$. Crop-truncation and referent-switch
categories are each $0.00\%$ on all three and omitted from the columns.
$^\dagger$Fraction of $[.9,1]$-baseline rows that fall to $[.75,.9)$ after
LFPR (the same transition-matrix cell Figure~\ref{fig:transition-matrix}
plots for Ref-L4/Flickr), computed per dataset. Zero new inference; reuses
the same per-row predictions as
Table~\ref{tab:app-e1-per-dataset}.}
\label{tab:app-per-dataset-taxonomy}
\end{table}

\begin{table*}[t]
\centering
\small
\resizebox{\linewidth}{!}{%
\begin{tabular}{lrrrrl}
\toprule
Arm & Acc@0.5 & Acc@0.75 & Acc@0.9 & mAcc & IoU W/L/T vs incumbent \\
\midrule
$B$ incumbent & 89.212 & 80.790 & 60.955 & 75.925 & --- \\
$R$ routing & 89.903 & 81.597 & 62.117 & 76.786 & 1{,}444 / 651 / 28{,}874 \\
$C$ crop & 89.448 & 81.462 & 62.799 & 76.723 & 15{,}725 / 10{,}608 / 4{,}636 \\
$CR^\dagger$ unguarded negative control & 86.974 & 77.632 & 56.382 & 72.707 & 11{,}683 / 16{,}010 / 3{,}276 \\
$CRG$ full (LFPR) & \textbf{90.029} & 81.643 & \textbf{60.925} & \textbf{76.570} & 13{,}925 / 12{,}357 / 4{,}687 \\
\bottomrule
\end{tabular}%
}
\caption{Five-arm same-row decomposition on the 30,969-row transfer split,
all against the same incumbent with 10,000-iteration image-cluster bootstrap
and Holm correction.  Every arm is significant at adjusted
$p\mathrel{\approx}0.003$ or better.  Routing changes few rows (2,095 of
30,969) but nearly always improves them.  \textbf{$\dagger$: $CR$ here is a
genuine unguarded negative control}, recomputed directly from $CRG$'s own
stored per-row fields (incumbent box, crop candidate, guard decision) with
the guard gate removed and every row unconditionally replaced -- it uses the
exact same crop candidates $CRG$ uses, so the comparison is a true ablation
of the guard alone.  It underperforms the incumbent on every metric,
confirming the guard is load-bearing rather than a trade against a
competitive alternative.  An archived, independently-generated $CR$ run
(Acc@0.5/0.75/0.9/mAcc $=89.919/82.311/64.604/77.727$) exists under the same
label; that artifact uses an identical guard/fallback/fusion mechanism to
$CRG$ (verified: its per-row admission decision reproduces $CRG$'s frozen
guard thresholds on 30,969/30,969 rows, 100.00\% exact) but different
upstream baseline and crop-candidate predictions from a separate execution,
so it was never a guard ablation at all.  That artifact is retained only as
provenance of the mislabeling, not as a result.}
\label{tab:app-e3-arms}
\end{table*}

\subsubsection{Component decomposition on identical rows}

Because the crop candidate is retained before the guards and the midpoint are
applied, the crop, guard, and fusion stages separate on the same 30,969 rows
without any additional inference.  Table~\ref{tab:app-components} reports the
paired contrasts.

\begin{table}[t]
\centering
\small
\begin{tabular}{lrr}
\toprule
Contrast and metric & $\Delta$ (pp) & 95\% CI \\
\midrule
\multicolumn{3}{l}{\emph{Resolution routing minus incumbent}}\\
\quad Acc@0.5 & $+0.691$ & $[0.554, 0.836]$ \\
\quad Acc@0.75 & $+0.807$ & $[0.637, 0.984]$ \\
\quad Acc@0.9 & $+1.162$ & $[0.927, 1.402]$ \\
\midrule
\multicolumn{3}{l}{\emph{Crop alone (guarded) minus incumbent}}\\
\quad Acc@0.5 & $+0.236$ & $[0.125, 0.356]$ \\
\quad Acc@0.75 & $+0.672$ & $[0.402, 0.940]$ \\
\quad Acc@0.9 & $+1.844$ & $[1.349, 2.350]$ \\
\midrule
\multicolumn{3}{l}{\emph{Unguarded crop+routing minus incumbent (negative
control)}}\\
\quad Acc@0.5 & $-2.926$ & $[-3.264, -2.587]$ \\
\quad Acc@0.75 & $-3.969$ & $[-4.461, -3.477]$ \\
\quad Acc@0.9 & $-5.748$ & $[-6.494, -5.002]$ \\
\midrule
\multicolumn{3}{l}{\emph{Full system (LFPR) minus incumbent}}\\
\quad Acc@0.5 & $+0.817$ & $[0.639, 1.001]$ \\
\quad Acc@0.75 & $+0.852$ & $[0.566, 1.147]$ \\
\quad Acc@0.9 & $-0.029$ & $[-0.554, 0.497]$ \\
\bottomrule
\end{tabular}
\caption{Same-row component decomposition, 10,000 image-cluster bootstrap
resamples, every contrast against the same incumbent.  ``Crop alone'' keeps
the frozen guard active (guard and routing are independent axes; this row
isolates the crop-only pipeline path, not a guard ablation).  The
``unguarded crop+routing'' row is a true guard ablation, recomputed directly
from $CRG$'s own stored candidate and guard fields with the guard gate
removed: it underperforms the incumbent on every metric, not merely on
Acc@0.5, so cropping does not buy strict-IoU accuracy for free once the
guard stops filtering bad candidates.  The guards and midpoint in the full
system then return a small further Acc@0.9 concession in exchange for the
largest Acc@0.5 gain.  All contrasts are significant at Holm-adjusted
$p\mathrel{\approx}0.003$ or better except the full system at Acc@0.9
($p=0.93$).}
\label{tab:app-components}
\end{table}

The routing contrast supersedes an earlier version of this table that reported
exact zeros for every routing metric.  The router flagged the same 2,223 of
30,969 rows (7.18\%) in both runs, but in the earlier one the requested pixel
floor was silently discarded by the processor, so the routed arm reproduced
the default-resolution arm byte for byte and every bootstrap replicate was
identically zero.  With the floor verified in effect the router changes 2,213
of those rows.

\subsubsection{Released-specialist comparison}

Two released referring-expression specialists were evaluated zero-shot on the
identical rows, clusters, and evaluator.  Paired contrasts against the frozen
incumbent are $+1.511$ points at Acc@0.5 for the 4B specialist
($[1.029, 2.002]$) and $+2.099$ for the 8B specialist ($[1.663, 2.508]$); at
Acc@0.9 the same contrasts are $+0.604$ ($[-0.317, 1.538]$, not
distinguishable from zero) and $-7.168$ ($[-8.178, -6.185]$).  In mean accuracy
the 4B specialist gains $+0.798$ points and the 8B specialist loses $1.220$.
Paired IoU outcomes against the incumbent are 15,318 wins / 15,311 losses /
340 ties for the 4B specialist and 11,428 wins / 19,195 losses / 346 ties for
the 8B specialist.

The deployed guarded arm (LFPR/$CRG$) trails the 4B specialist by $-0.694$
points at Acc@0.5 ($[-1.148, -0.234]$) and is statistically indistinguishable
from it at Acc@0.9 ($-0.633$, $[-1.545, 0.228]$); against the 8B specialist it
trails by $-1.282$ at Acc@0.5 ($[-1.698, -0.883]$) and leads by $+7.139$ at
Acc@0.9 ($[6.169, 8.125]$).  Scaling the specialist from 4B to 8B gives
$+0.588$ points at Acc@0.5 ($[0.171, 1.011]$), $-7.772$ at Acc@0.9
($[-8.816, -6.730]$), and $-2.017$ in mean accuracy.

As a negative control, the same comparison was run against the genuine
unguarded arm from Table~\ref{tab:app-e3-arms} (guard gate removed, same
underlying candidates as $CRG$).  It trails both specialists at Acc@0.5 by a
wider margin than the guarded arm ($-3.749$ against the 4B specialist,
$-4.337$ against the 8B; point estimates, since the specialists were not
independently re-scored against this recomputed arm) and no longer leads
either specialist convincingly at Acc@0.9: $-5.176$ against the 4B specialist
and $+2.596$ against the 8B, versus the guarded arm's $-0.633$ and $+7.139$
respectively.  Removing the guard therefore does not trade Acc@0.5 for a
stronger strict-IoU position relative to the specialists; it loses ground
at Acc@0.5 without gaining a comparably strong position at Acc@0.9.

Two controls bound this comparison.  An image-level firewall audit compared
17,978 unique training images from the specialists' published training
manifest against the 3,982 unique evaluation images, partition by partition,
and found zero overlap in all five partitions.  This rules out exact-image
memorization only; it does not rule out semantic overlap, near-duplicate
images, shared class priors, or shared annotation style between the
specialists' training data and our evaluation split, any of which could
still inflate a specialist's advantage over a model that never trained on
this distribution at all.  We narrow the claim accordingly: the firewall
supports a claim of no exact-image leakage, not a claim that the specialist
comparison isolates model quality from training-distribution similarity.
Separately, because midpoint
fusion produces continuous coordinates while all three single-pass systems
emit coordinates on a $0.001$ grid, we requantized our predictions to that
grid and rescored.  Every metric moved by less than $0.03$ points, so
coordinate granularity accounts for none of the differences reported here.
The quantization check was run against the superseded predictions; because it
bounds a rounding effect rather than a treatment effect, and because the
re-measured arms use the identical fusion arithmetic, the bound carries over
unchanged.

\subsection{Flickr30K Entities standard-protocol (merged-box) confirmation}
\label{app:flickr-merged}

The single-box subset used in the main text excludes every multi-box phrase,
so it is not comparable to standard Flickr30K Entities phrase-grounding
numbers, which merge a phrase's boxes into one enclosing target rather than
discarding it.  We therefore ran a second, separately-frozen confirmation
that keeps every visible entity phrase with at least one annotated box --
14,481 rows over 999 images (11,574 single-box plus 2,907 multi-box phrases,
each represented by the deterministic enclosing $xyxy$ box of its annotated
boxes) -- with its own denominator, image-hash manifest, and access ledger
kept separate from the single-box subset so the two protocols cannot be
confused.  An image-level firewall audit against the historical Ref-L4 test
and validation splits found zero overlapping images and zero overlapping
record keys.

Full intervals: Acc@0.5 rises $+0.815$ points (95\% CI $[0.299, 1.324]$,
Holm-adjusted $p=0.0052$), Acc@0.75 rises $+1.326$ ($[0.714, 1.933]$,
$p=0.0008$), Acc@0.9 rises $+1.022$ ($[0.322, 1.705]$, $p=0.0052$), and mAcc
rises $+0.973$ ($[0.648, 1.310]$, $p=0.0008$), with zero invalid predictions
in either arm.  The gain pattern differs from the single-box result: there
Acc@0.75 and Acc@0.9 lead Acc@0.5; here all four primary metrics move
together by roughly the same amount.  Taken together with the single-box
result, LFPR improves both the historical single-box selection and the
standard merged-box protocol on the same held-out test images, so the
single-box result is not an artifact of excluding multi-box phrases.

\subsection{Mechanism replication on the Flickr30K merged-box confirmation}
\label{app:flickr-mechanism}

The transfer-split same-row decomposition (Table~\ref{tab:app-e3-arms},
Table~\ref{tab:app-components}) is the paper's most detailed causal
evidence for the guard and router on frozen transfer data, but it shares
the RefCOCO-family evaluation used above.  We therefore repeated the
identical $B/R/C/CR^\dagger/CRG$ decomposition on the Flickr30K merged-box
confirmation above, which never informed method development, reconstructing
every arm from the frozen run's own archived per-row fields with no
additional model inference.

\begin{table}[t]
\centering
\small
\resizebox{\linewidth}{!}{%
\begin{tabular}{lrrrrr}
\toprule
Arm & Acc@0.5 & Acc@0.75 & Acc@0.9 & mAcc & Mean IoU $\Delta$ vs $B$ \\
\midrule
$B$ incumbent (pre-resolution) & 74.864 & 59.720 & 37.753 & 56.975 & --- \\
$R$ routing alone & 75.893 & 61.784 & 39.749 & 58.585 & +1.013 \\
$C$ guarded crop alone & 74.926 & 59.050 & 37.200 & 56.463 & $-0.111$ \\
$CR^\dagger$ unguarded negative control & 74.353 & 58.484 & 36.662 & 55.952 & $-0.652$ \\
$CRG$ full (LFPR) & \textbf{75.678} & \textbf{61.046} & \textbf{38.775} & \textbf{57.948} & \textbf{+0.712} \\
\bottomrule
\end{tabular}%
}
\caption{Same-row $B/R/C/CR^\dagger/CRG$ decomposition on the Flickr30K
merged-box confirmation (14,481 rows, 999 images), reconstructed from the
frozen run's own archived per-row fields with no new model inference.
Deltas are against $B$, the pre-resolution incumbent, matching
Table~\ref{tab:app-e3-arms}'s convention.  10,000-iteration image-cluster
bootstrap.}
\label{tab:app-flickr-mechanism}
\end{table}

The two unconditional resolution controls from P2-2 were also scored on the
same full 14,481-row population, rather than only on the 512-row matched-cost
subset.  Their absolute scores and paired mean-IoU changes against the same
pre-resolution incumbent are shown in Table~\ref{tab:app-flickr-unconditional-full}.

% Generated from metrics_canonical.csv by generate_latex_tables.py (see the
% supplementary code, metrics_artifacts/); inlined here so this submission
% package has no external file dependency.
\begin{table}[t]
\centering
\small
\resizebox{\linewidth}{!}{%
\begin{tabular}{lrrrrr}
\toprule
Arm & Acc@0.5 & Acc@0.75 & Acc@0.9 & mAcc & Mean IoU $\Delta$ vs $B$ \\
\midrule
$B$ incumbent (pre-resolution) & 74.864 & 59.720 & 37.753 & 56.975 & --- \\
$R_{\mathrm{ALL}}$ unconditional floor & 74.629 & 61.398 & 42.469 & 58.747 & +0.692 \\
$CR_{\mathrm{ALL}}$ unconditional floor+crop & 74.774 & 61.149 & 41.551 & 58.461 & +0.591 \\
\bottomrule
\end{tabular}%
}
\caption{Full-population P2-2 unconditional controls on the Flickr30K
merged-box confirmation (14,481 rows, 999 images). Mean-IoU deltas
are paired against the pre-resolution incumbent $B$; no new inference
was performed for the $B$ row.}
\label{tab:app-flickr-unconditional-full}
\end{table}

The pattern matches the Ref-L4 and RefCOCO-family decompositions: $CRG$
beats $B$ significantly at both the permissive and strict threshold (95\%
CI $[+0.293,+1.338]$ points at Acc@0.5, $[+0.319,+1.698]$ at Acc@0.9), and
the guard remains load-bearing -- $CR^\dagger$ is not merely worse than
$CRG$, it is significantly \emph{below} $B$ at Acc@0.9 (CI
$[-1.884,-0.299]$), the same asymmetry reported for Ref-L4 and the RefCOCO
family.  The router's own isolated contribution is positive and
significant at every threshold (Acc@0.5 CI $[+0.526,+1.536]$, Acc@0.9 CI
$[+1.417,+2.575]$) and larger in magnitude than on the RefCOCO family; this
is not a new mechanism, it tracks a higher routing rate on this population
-- the router changes the predicted box on 4,318 of 14,481 rows (29.8\%)
here, against 2,213 of 30,969 (7.1\%) on the RefCOCO family
(Section~\ref{sec:transfer}), consistent with Flickr30K Entities' generally
smaller annotated targets.

The same guard-by-fusion $2\times2$ factorial reported for Ref-L4
(Table~\ref{tab:app-m2-factorial}) is computable on this dataset with the
identical zero-new-inference method: the incumbent box, crop candidate,
and guard decision are already stored per row.

\begin{figure*}[t]
\centering
\small
\setlength{\tabcolsep}{4pt}
\resizebox{\linewidth}{!}{%
\begin{tabular}{cc}
\begin{tabular}{c|cc}
\multicolumn{3}{c}{\textbf{Ref-L4}} \\
& \textbf{Midpoint} & \textbf{Replacement} \\
\hline
\textbf{Guarded} & \shortstack{Acc@0.5: 89.725\\Acc@0.9: 61.142\\mAcc: 76.013\\Mean IoU: +0.691} & \shortstack{Acc@0.5: 89.399\\Acc@0.9: 63.143\\mAcc: 76.166\\Mean IoU: +0.712} \\
\textbf{Unguarded} & \shortstack{Acc@0.5: 88.262\\Acc@0.9: 59.350\\mAcc: 73.864\\Mean IoU: $-0.454$} & \shortstack{Acc@0.5: 87.247\\Acc@0.9: 62.059\\mAcc: 74.491\\Mean IoU: $-0.884$} \\
\end{tabular}
&
\begin{tabular}{c|cc}
\multicolumn{3}{c}{\textbf{Flickr30K merged-box}} \\
& \textbf{Midpoint} & \textbf{Replacement} \\
\hline
\textbf{Guarded} & \shortstack{Acc@0.5: 75.678\\Acc@0.9: 38.775\\mAcc: 57.948\\Mean IoU: $-0.301$} & \shortstack{Acc@0.5: 74.926\\Acc@0.9: 37.200\\mAcc: 56.463\\Mean IoU: $-1.123$} \\
\textbf{Unguarded} & \shortstack{Acc@0.5: 75.057\\Acc@0.9: 38.071\\mAcc: 57.003\\Mean IoU: $-0.726$} & \shortstack{Acc@0.5: 74.353\\Acc@0.9: 36.662\\mAcc: 55.952\\Mean IoU: $-1.665$} \\
\end{tabular}
\end{tabular}%
}
\caption{Guard $\times$ fusion factorial on the retrospective Ref-L4 split and
prospective Flickr30K merged-box confirmation.  The guarded-midpoint cell
($CRG$) is the shipped policy in both cases; the unguarded cells lose
accuracy on every reported threshold, showing that the guard, not the fusion
rule alone, is the load-bearing component.}
\label{fig:guard-fusion-factorial}
\end{figure*}

\begin{table*}[t]
\centering
\small
\resizebox{\linewidth}{!}{%
\begin{tabular}{lrrrrrl}
\toprule
Cell & Acc@0.5 & Acc@0.75 & Acc@0.9 & mAcc & Mean IoU $\Delta$ & IoU W/L/T \\
\midrule
Guarded, midpoint ($CRG$, shipped) & \textbf{75.678} & 61.046 & 38.775 & 57.948 & $-0.301$ & 5{,}024 / 5{,}471 / 3{,}986 \\
Guarded, replacement & 74.926 & 59.050 & 37.200 & 56.463 & $-1.123$ & 4{,}430 / 6{,}075 / 3{,}976 \\
Unguarded, midpoint & 75.057 & 59.747 & 38.071 & 57.003 & $-0.726$ & 5{,}541 / 6{,}003 / 2{,}937 \\
Unguarded, replacement ($CR^\dagger$) & 74.353 & 58.484 & 36.662 & 55.952 & $-1.665$ & 4{,}876 / 6{,}686 / 2{,}919 \\
\bottomrule
\end{tabular}%
}
\caption{Guard $\times$ fusion factorial on the Flickr30K merged-box
confirmation (14,481 rows, 13,360 with an admitted crop candidate;
10,000-iteration image-cluster bootstrap per cell, no new inference). Mean
IoU $\Delta$ is against the post-routing incumbent $R$ (this factorial's
own baseline), not against $B$ as in Table~\ref{tab:app-flickr-mechanism}
above -- the two tables use different reference arms by construction, so
their $\Delta$ columns are not directly comparable. The same pattern as
Ref-L4 holds: the guard, not the fusion rule, is what
protects accuracy here -- both guarded cells beat both unguarded cells at
every threshold, and only the unguarded cells fall below the shipped
policy. Unlike Ref-L4, guarded replacement does not trade Acc@0.5 for a
higher Acc@0.9 on this dataset ($37.200$ versus $CRG$'s $38.775$); on
Flickr the guarded-midpoint cell dominates guarded-replacement on every
metric, so the trade Ref-L4 exhibits is not a universal property of the
replacement rule.}
\label{tab:app-flickr-factorial}
\end{table*}

\begin{figure}[t]
\centering
\includegraphics[width=0.85\linewidth]{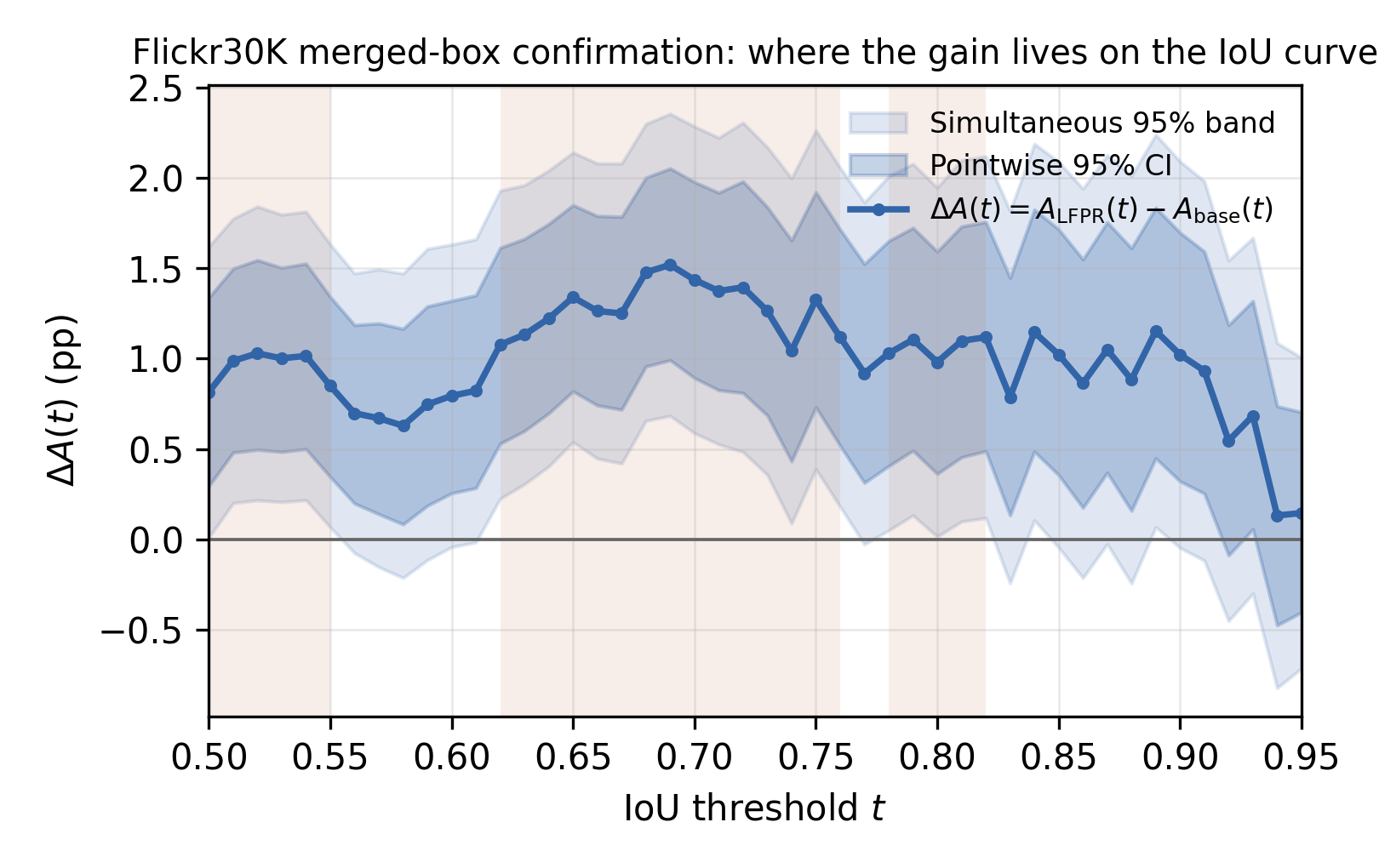}
\caption{Treatment-minus-incumbent accuracy, $\Delta A(t)=A_{\mathrm{LFPR}}(t)-A_{\mathrm{base}}(t)$,
as a function of the IoU threshold on the Flickr30K merged-box confirmation
(14,481 rows, 999 images).  The darker band is the pointwise 95\% interval;
the lighter band is the simultaneous (max-$|T|$) 95\% band used to license
concentration claims without cherry-picking a threshold.  Shaded vertical
strips mark the grid points where the simultaneous band excludes zero.  The
gain is positive across nearly the entire threshold range rather than
concentrated at one operating point.}
\label{fig:flickr-iou-curve}
\end{figure}

A dense-grid extension of this comparison (46 thresholds,
$t=0.50,0.51,\ldots,0.95$, one shared cluster-bootstrap resample per
iteration across the whole grid, and a simultaneous max-$|T|$ band rather
than a pointwise one) finds the $CRG$-versus-$B$ gain simultaneously
significant on $t\in[0.50,0.55]\cup[0.62,0.76]\cup[0.78,0.82]$ plus two
isolated grid points at $0.84$ and $0.89$: the effect is not concentrated
at one threshold, it holds across most of the permissive-to-strict range on
this prospective evidence.  Figure~\ref{fig:flickr-iou-curve} plots the full
curve.  Appendix~\ref{app:all-datasets-curve} extends this same dense-grid
construction to all five evaluations using each evaluation's own
pre-resolution and fused per-row prediction files.

Table~\ref{tab:app-flickr-cost} extends the same synchronized cost audit
already reported for the RefCOCO-family split (Table~\ref{tab:app-cost-r3})
to the Flickr30K merged-box confirmation, using the identical protocol (one
untimed warm-up, three CUDA-synchronized timed repeats per row, same
RTX~3090). This closes the gap the previous revision cycle left open: $B$/
$R$/$C$/$CR$/$CRG$ were timed on this dataset for the first time only after
the unconditional controls, and the five guarded arms' numbers were computed
but never written up as a table. Zero runtime errors and zero invalid
outputs on every arm and size.

\begin{table}[t]
\centering
\small
\setlength{\tabcolsep}{3.5pt}
\begin{tabular}{lrrrrrr}
\toprule
Rows & Arm & Calls & Mean s & P95 s & Max alloc.\ GiB & Invalid \% \\
\midrule
128 & $B$ & 1.000 & 0.837 & 0.900 & 16.44 & 0.00 \\
128 & $R$ & 1.156 & 1.299 & 3.847 & 17.58 & 0.00 \\
128 & $C$ & 2.000 & 1.646 & 1.821 & 16.44 & 0.00 \\
128 & $CR$ & 2.156 & 2.160 & 4.717 & 17.58 & 0.00 \\
128 & $CRG$ & 2.156 & 2.244 & 4.734 & 17.58 & 0.00 \\
128 & $R_{\mathrm{ALL}}$ & 1.000 & 3.072 & 3.225 & 17.59 & 0.00 \\
128 & $CR_{\mathrm{ALL}}$ & 2.000 & 5.928 & 6.023 & 17.60 & 0.00 \\
512 & $B$ & 1.000 & 0.878 & 0.982 & 16.44 & 0.00 \\
512 & $R$ & 1.250 & 1.671 & 4.057 & 17.58 & 0.00 \\
512 & $C$ & 2.000 & 1.813 & 2.277 & 16.44 & 0.00 \\
512 & $CR$ & 2.250 & 2.515 & 4.895 & 17.58 & 0.00 \\
512 & $CRG$ & 2.250 & 2.417 & 4.693 & 17.58 & 0.00 \\
512 & $R_{\mathrm{ALL}}$ & 1.000 & 3.003 & 3.075 & 17.59 & 0.00 \\
512 & $CR_{\mathrm{ALL}}$ & 2.000 & 6.103 & 6.489 & 17.60 & 0.00 \\
\bottomrule
\end{tabular}
\caption{Synchronized cost audit on the Flickr30K merged-box confirmation, all
seven arms, same protocol and same frozen 128/512-row prefixes as
Table~\ref{tab:app-cost-r3}. $CR$ and $CRG$ share the identical call budget
and generation calls (they differ only in which box the already-generated
crop candidate is turned into -- hard replacement versus midpoint fusion --
a CPU-only decision that costs nothing), so their latencies match within
timing noise, mirroring the same pattern already reported for the
RefCOCO-family audit. $R$'s calls exceed 1.000 by exactly its label-free
routing rate on this row prefix (e.g.\ $0.250$ at 512 rows, i.e.\ $25\%$),
consistent with the $29.8\%$ population-level routing rate reported in
Appendix~\ref{app:flickr-mechanism} -- the two figures differ because this
is a smaller 512-row sample, not a second measurement of the same
population. Peak allocated memory is $16.44$~GiB for the two arms that
never raise resolution ($B$, $C$) and $17.58$--$17.60$~GiB for every arm
that can trigger the resolution floor.}
\label{tab:app-flickr-cost}
\end{table}

Unlike the RefCOCO-family measurement, $R_{\mathrm{ALL}}$/$CR_{\mathrm{ALL}}$
accuracy was also scored on Flickr, on the same frozen 512-row subset used
for the cost audit above (single pass per row, no new inference beyond what
the cost audit already ran).  A matched-denominator readout for the other
five arms was not available in the previous revision cycle -- $B$/$R$/$C$/
$CR^\dagger$/$CRG$ had only been scored on the full 14,481-row population --
so the two groups could only be compared qualitatively.  We closed that gap
by restricting the archived full-population $B$ and crop-pass artifacts to
the identical 512 rows (row selection matched by \texttt{record\_key}
against the cost audit's own frozen manifest, zero new model inference) and
rescoring all seven arms on that one matched set, reported in
Table~\ref{tab:app-flickr-matched512}.

\begin{table}[t]
\centering
\small
\resizebox{\linewidth}{!}{%
\begin{tabular}{lrrrrr}
\toprule
Arm & Acc@0.5 & Acc@0.75 & Acc@0.9 & mAcc & Mean IoU $\Delta$ vs $B$ \\
\midrule
$B$ incumbent (pre-resolution) & 74.219 & 64.648 & 42.773 & 59.648 & --- \\
$R$ routing alone & 75.586 & 66.602 & 44.531 & 61.387 & $+1.175$ \\
$C$ guarded crop alone & 74.805 & 63.086 & 42.578 & 59.668 & $+0.323$ \\
$CR^\dagger$ unguarded negative control & 74.805 & 63.086 & 41.992 & 59.609 & $-0.082$ \\
$CRG$ full (LFPR) & 75.000 & 66.406 & 43.555 & 60.723 & $+0.932$ \\
$R_{\mathrm{ALL}}$ unconditional floor & 75.586 & 67.773 & 48.047 & 62.617 & $+1.828$ \\
$CR_{\mathrm{ALL}}$ unconditional floor+crop & 75.391 & 67.383 & 45.117 & 61.816 & $+1.590$ \\
\bottomrule
\end{tabular}%
}
\caption{All seven arms scored on the identical frozen 512-row Flickr30K
merged-box subset (36 images) used for the synchronized cost audit
(Table~\ref{tab:app-flickr-cost}), so accuracy and cost are now
comparable on the same denominator throughout this section.  $B$/$R$/$C$/
$CR^\dagger$/$CRG$ are restricted from the archived full-population run
(Table~\ref{tab:app-flickr-mechanism}) by exact \texttt{record\_key} match,
not rerun; $R_{\mathrm{ALL}}$/$CR_{\mathrm{ALL}}$ are the same predictions
already used for the cost audit.  Deltas and the 10,000-iteration
image-cluster bootstrap are against this table's own 512-row $B$, not the
14,481-row $B$ used elsewhere in this section.}
\label{tab:app-flickr-matched512}
\end{table}

The point estimates reproduce the qualitative pattern already suggested by
the unmatched comparison -- every treatment arm sits numerically above $B$,
and the two unconditional controls sit above the four selective arms on
Acc@0.9 and mAcc -- but at 512 rows drawn from only 36 images, the
image-cluster bootstrap resolves very little of it: only $R$'s mAcc gain
(95\% CI $[+0.336,+3.374]$) and $R_{\mathrm{ALL}}$'s mAcc gain (CI
$[+0.116,+5.776]$) exclude zero; every other arm's interval, including
$CRG$'s (CI $[-0.427,+2.749]$ at mAcc) and $CR_{\mathrm{ALL}}$'s (CI
$[-0.591,+4.903]$ at mAcc), does not.  This is a sample-size limitation of
the 512-row subset, not a reversal of the full-population finding: the same
$CRG$-versus-$B$ contrast is significant at every threshold on the full
14,481-row population (Table~\ref{tab:app-flickr-mechanism}).  We therefore
report Table~\ref{tab:app-flickr-matched512} as what it is -- a
matched-denominator accuracy-cost readout, not an additional significance
result -- and continue to rely on the full-population table for the paper's
confirmatory claims.  Within this matched set, $CR_{\mathrm{ALL}}$ does not
improve on $R_{\mathrm{ALL}}$ despite costing $2\times$ the calls: the
additional unconditional crop pass is a net negative on every metric at
this sample size, reinforcing why the deployed policy routes crop attempts
selectively rather than applying them everywhere.

\begin{figure}[t]
\centering
\includegraphics[width=\linewidth]{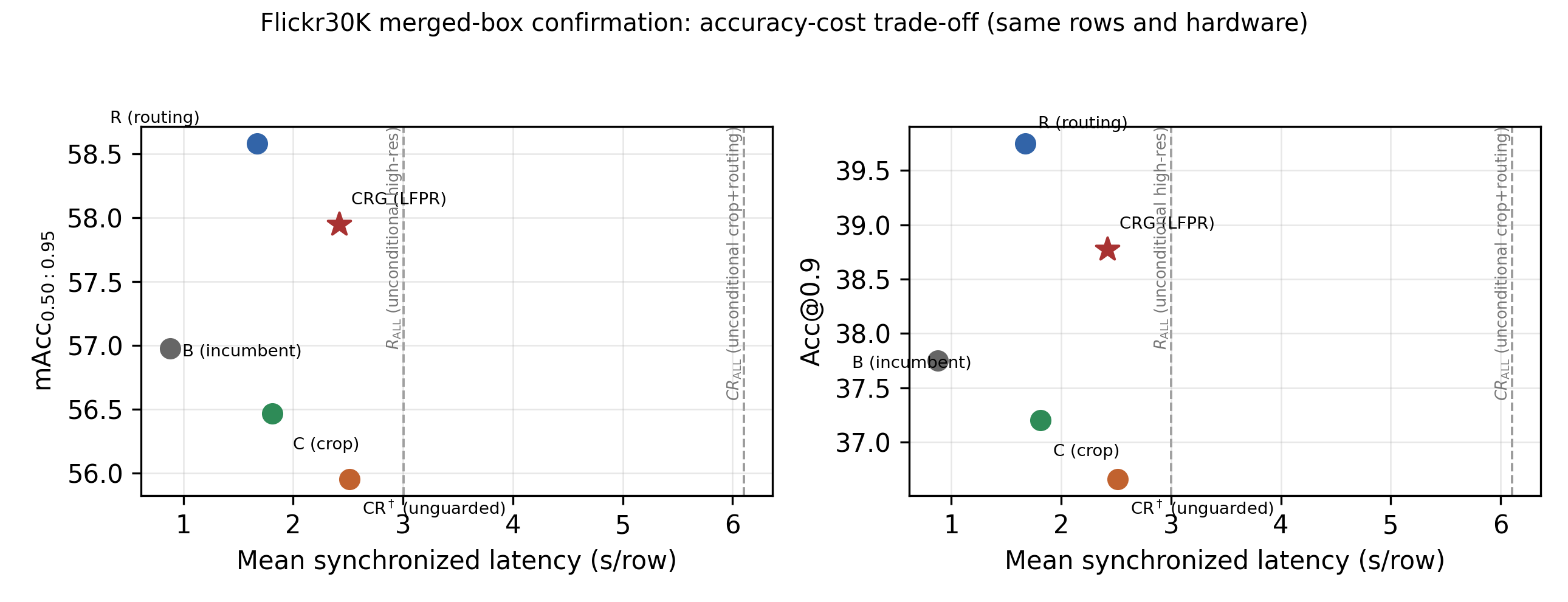}
\caption{\textbf{Accuracy-cost trade-off on the Flickr30K merged-box
confirmation.} Each point uses the same 14,481-row evaluation (accuracy,
Table~\ref{tab:app-flickr-mechanism}) and the same synchronized 512-row
cost prefix on the same RTX~3090 (cost, this section); no new inference
was run to build this figure. $R_{\mathrm{ALL}}$/$CR_{\mathrm{ALL}}$
(dashed reference lines) cost $3.42\times$/$6.95\times$ $B$; their own
matched-sample accuracy is reported in the surrounding text rather than
plotted here, since it uses the smaller 512-row sample rather than the
full population the other five points use. On this dataset, $R$ alone
(not $CRG$) is the accuracy-maximizing point on the frontier -- see text.}
\label{fig:flickr-pareto}
\end{figure}

Figure~\ref{fig:flickr-pareto} plots the accuracy-cost trade-off for this
dataset, pairing each of the five $B/R/C/CR^\dagger/CRG$ arms' own
full-population accuracy (Table~\ref{tab:app-flickr-mechanism}) with its
own synchronized 512-row cost measurement above; no new inference was run
to build this figure. Unlike the RefCOCO-family panel
(Figure~\ref{fig:refcoco-pareto}), where $CRG$ is the best matched-cost
mAcc point, here $R$ alone reaches the highest mAcc and Acc@0.9 of any
arm, including $CRG$, at lower cost ($1.671$ vs.\ $2.417$s/row) -- the
crop/guard/fusion component contributes nothing beyond routing at the
mAcc level on this dataset, consistent with the router's outsized
isolated contribution already noted above (a $29.8\%$ routing rate here
against $7.1\%$ on the RefCOCO family). $CRG$ still beats $B$
significantly at every threshold (Table~\ref{tab:app-flickr-mechanism})
and the guard remains necessary -- $CR^\dagger$ falls to the lowest mAcc
and Acc@0.9 of any arm here -- but on Flickr the accuracy-maximizing point
on this frontier is $R$, not the full shipped policy. This is reported
for completeness rather than folded into the main claim: the shipped
policy is fixed across every evaluation in this paper, and this
trade-off reflects Flickr's own routing rate, not evidence for changing
the deployed rule.

\subsection{IoU gain curve across all five evaluations}
\label{app:all-datasets-curve}

\begin{figure}[t]
\centering
\includegraphics[width=0.85\linewidth]{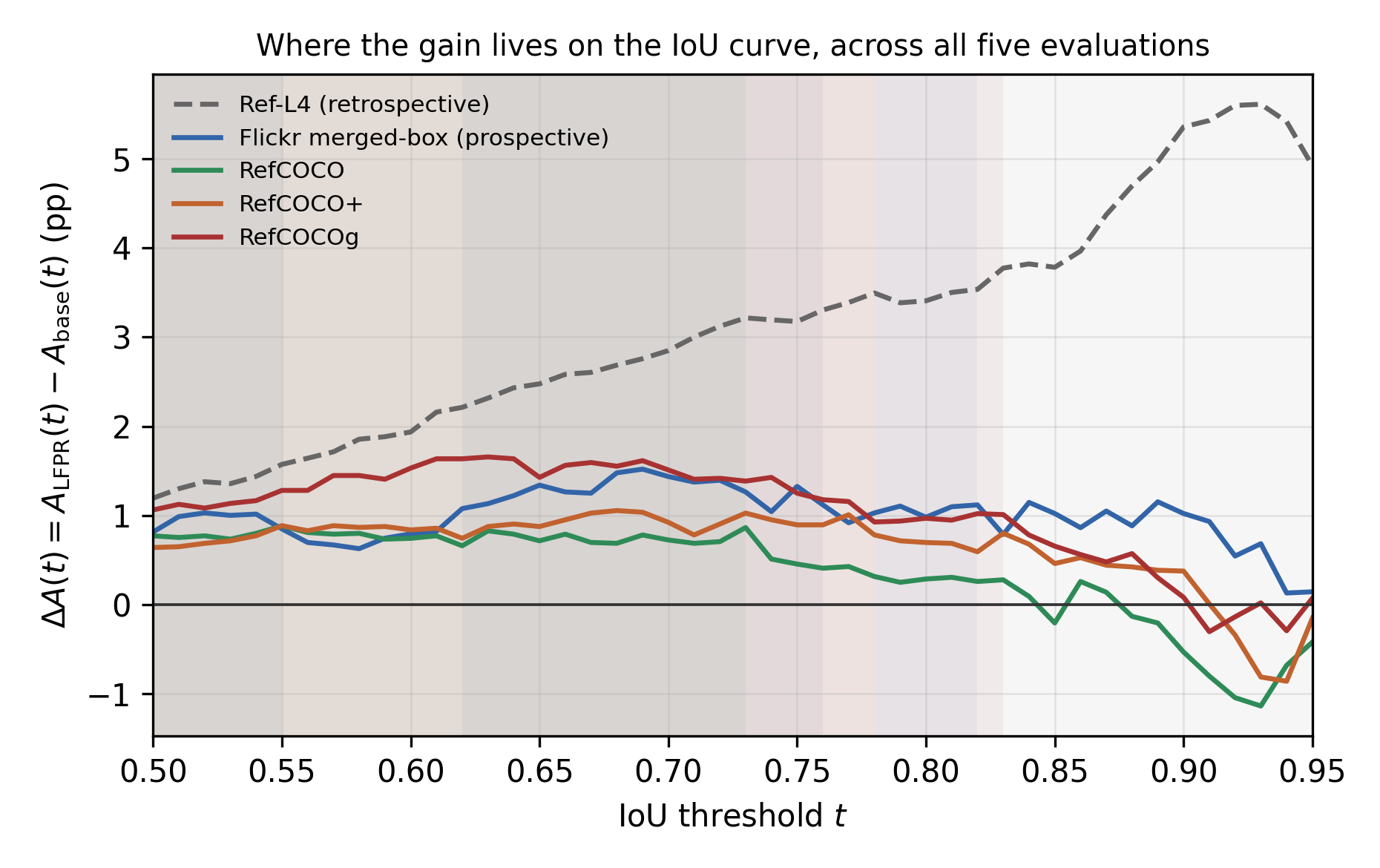}
\caption{Treatment-minus-incumbent accuracy, $\Delta A(t)$, as a function
of the IoU threshold, across all five evaluations, each against its own
true pre-resolution incumbent $B$. Shaded vertical strips mark each
series' own simultaneous-band significant intervals. Ref-L4 and Flickr
(retrospective and prospective evidence respectively) grow larger at
strict thresholds; RefCOCO/RefCOCO+/RefCOCOg (shown individually, not
pooled) stay positive through most of the range and turn negative only
close to $t=0.85$--$0.92$, consistent with the deployed policy's broad
per-dataset improvement and softening strict-IoU effect reported at fixed
thresholds.}
\label{fig:all-datasets-curve}
\end{figure}

Figure~\ref{fig:flickr-iou-curve} above shows the dense-grid curve for the
Flickr30K merged-box confirmation alone. Figure~\ref{fig:all-datasets-curve}
overlays the same $\Delta A(t)=A_{\mathrm{LFPR}}(t)-A_{\mathrm{base}}(t)$
curve, computed with the identical protocol (pointwise and simultaneous
max-$|T|$ image-cluster bootstrap, 10,000 iterations) and against each
evaluation's own true pre-resolution incumbent $B$, for all five
evaluations: Ref-L4 (retrospective), Flickr30K merged-box (prospective),
and RefCOCO/RefCOCO+/RefCOCOg shown individually rather than pooled. Each
curve is built from a separately-verified $B$/$CRG$ per-row pair -- Ref-L4
from \texttt{baseline\_predictions.jsonl} and \texttt{fused.jsonl}, Flickr
from the merged-box confirmation's own \texttt{baseline/merged.jsonl} and
\texttt{lfpr\_predictions.jsonl}, and RefCOCO/RefCOCO+/RefCOCOg from the
true pre-resolution artifact used to correct Table~\ref{tab:app-e1-per-dataset}
-- and at $t\in\{0.50,0.75,0.90\}$ every curve reproduces its evaluation's
own fixed-threshold point estimate to within $10^{-10}$, the machine-checked
consistency gate this figure requires.

All five curves start positive at $t=0.50$ and stay positive through most of
the permissive-to-mid range.  Ref-L4 and Flickr grow \emph{larger} at strict
thresholds (Ref-L4 reaches $+5.35$ points at $t=0.90$ and stays above $+5$
through $t=0.94$); RefCOCO/RefCOCO+/RefCOCOg instead grow smaller and turn
negative only close to the strictest end of the range -- RefCOCO crosses
zero between $t=0.84$ and $0.85$ (reaching $-1.13$ at $t=0.93$), RefCOCO+
between $t=0.91$ and $0.92$ (reaching $-0.86$ at $t=0.94$), and RefCOCOg
between $t=0.90$ and $0.91$ (a shallow dip to $-0.30$ at $t=0.91$ before
recovering to slightly positive by $t=0.95$).  This is consistent with
every fixed-threshold result already reported: the deployed policy improves
Acc@0.5/0.75/mAcc/mean~IoU broadly across the RefCOCO family
(Table~\ref{tab:app-e1-per-dataset}), and its strict-IoU effect softens
toward the top of the threshold range rather than being uniformly positive
like Ref-L4 and Flickr's.

\subsection{Synchronized cost and robustness audit}
\label{app:cost-audit}

The generalist cost audit uses deterministic 128- and 512-row prefixes of the
frozen transfer order.  Every arm runs in a fresh process with batch size one,
one untimed warm-up, and three CUDA-synchronized timed repeats per row.  Both
sizes passed the declared admission checks for complete exact-key
manifests, finite timings, and absence of OOM or unclassified runtime
failures.  A later processor audit showed that the requested high-resolution
floor had not been applied in $R$, $CR$, or $CRG$; those timing values were
withdrawn.  The unaffected $B$ and $C$ measurements are retained below, and
Table~\ref{tab:app-cost-r3} restores $R$/$CR$/$CRG$ with the floor verified
in effect, alongside the two unconditional controls $R_\text{ALL}$ and
$CR_\text{ALL}$.

An earlier generalist timing table reported $B$/$C$ mean latency of
343.7/683.5~ms (128 rows) and 347.6/677.1~ms (512 rows).  That table is
withdrawn: the canonical policy-matched audit below, run under the same
declared protocol, measures $B$ at 0.945--1.049~s and $C$ at
1.876--2.315~s per row -- roughly $2.7\times$ higher.  We could not
reconcile the discrepancy from the archived environment record (neither
audit's log fixes the generation-token count, software commit, or GPU clock
state precisely enough to attribute the gap to a specific cause), so rather
than report two mutually inconsistent latency figures under the same label,
we retain only the canonical audit below as the paper's cost claim.

\begin{table}[t]
\centering
\small
\setlength{\tabcolsep}{3.5pt}
\begin{tabular}{lrrrrr}
\toprule
Arm & Calls & Mean s & P95 s & Max alloc.\ GiB & Invalid \% \\
\midrule
$B$ & 1.000 & 0.945 & 1.049 & 16.49 & 0.39 \\
$C$ & 1.996 & 1.876 & 2.315 & 16.49 & 0.39 \\
$R$ & 1.025 & 1.048 & 1.252 & 17.58 & 0.39 \\
$CR$ & 2.021 & 1.977 & 2.347 & 17.58 & 0.39 \\
$CRG$ & 2.021 & 1.865 & 1.985 & 17.58 & 0.39 \\
$R_\text{ALL}$ & 1.000 & 3.291 & 3.510 & 17.59 & 0.00 \\
$CR_\text{ALL}$ & 2.000 & 6.121 & 6.522 & 17.60 & 0.00 \\
\bottomrule
\end{tabular}
\caption{Canonical policy-matched 512-row synchronized cost audit with the
resolution floor verified in effect, including the unconditional
$R_\text{ALL}$/$CR_\text{ALL}$ controls.  One untimed warm-up plus three
CUDA-synchronized timed repeats per row, zero runtime errors on every arm.
This table supersedes the earlier $B$/$C$ figures withdrawn above; treat
only these absolute latencies, not the earlier ones, as the paper's cost
claim.
Selective routing ($R$/$CR$/$CRG$) costs little beyond $B$/$C$ because only
the flagged small-object rows pay the high-resolution premium; forcing that
premium onto every row ($R_\text{ALL}$/$CR_\text{ALL}$) costs
$3.10$--$3.14\times$ more than its selective counterpart at matched call
count, quantifying why selective routing is the deployed policy rather than
an always-high-resolution default.}
\label{tab:app-cost-r3}
\end{table}

\begin{figure}[t]
\centering
\includegraphics[width=\linewidth]{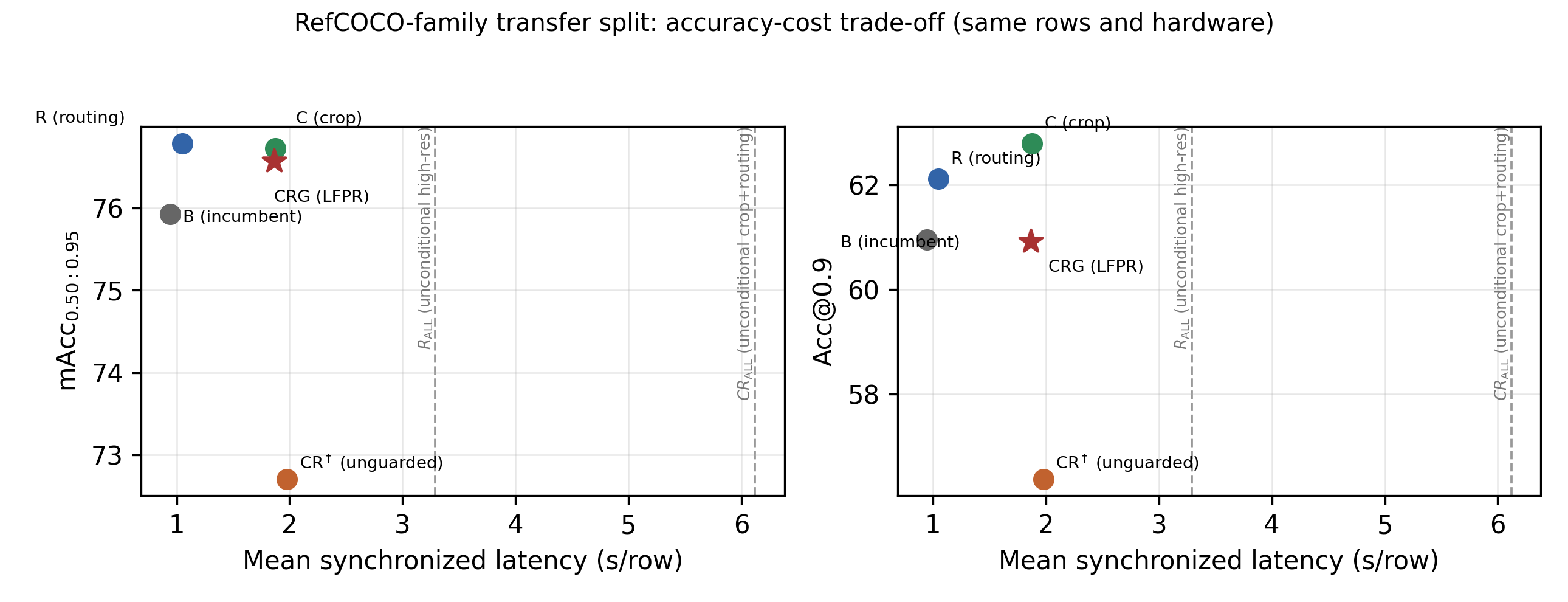}
\caption{\textbf{Accuracy-cost trade-off on the RefCOCO-family transfer
split.} Each point uses the same 30,969-row evaluation (accuracy,
Table~\ref{tab:app-e3-arms}) and the same synchronized 512-row cost prefix
on the same RTX~3090 (latency, Table~\ref{tab:app-cost-r3}); no new
inference was run to build this figure. $CRG$ (LFPR, starred) costs
$1.865$s/row against $B$'s $0.945$s/row, a $1.97\times$ increase --
consistent with the paper's ``approximately doubles inference latency''
characterization, since every admitted row pays one crop pass and
router-flagged rows may pay an additional high-resolution pass. The two
unconditional controls this cycle added, $R_{\mathrm{ALL}}$ and
$CR_{\mathrm{ALL}}$ (dashed reference lines), cost $3.48\times$ and
$6.48\times$ $B$ respectively -- $3.10$--$3.14\times$ more than their
selective counterparts $R$ and $CR^\dagger$ at matched call count
(Table~\ref{tab:app-cost-r3}) -- for a policy that no longer distinguishes
rows worth the extra pass from rows that are not. Their accuracy is now
measured on a matched 26,488-row RefCOCO validation split
(Table~\ref{tab:app-major-h-refcoco-val}): both beat the matched $B$ arm
across all reported metrics, while $R_{\mathrm{ALL}}$ is stricter-threshold
better than $CR_{\mathrm{ALL}}$ (Acc@0.9/mAcc). Because that accuracy readout
is on a validation split and this figure's selective-arm points remain on the
frozen test partitions, this panel is still a cost-vs-cost picture rather
than a same-split selective-versus-unconditional accuracy frontier.}
\label{fig:refcoco-pareto}
\end{figure}

Figure~\ref{fig:refcoco-pareto} plots the accuracy-cost trade-off implied by
Tables~\ref{tab:app-e3-arms} and~\ref{tab:app-cost-r3} directly, addressing
the question of why LFPR does not simply spend its second forward pass on
another unconditional high-resolution pass.

\begin{table}[t]
\centering
\small
\resizebox{\linewidth}{!}{%
\begin{tabular}{lrrrrr}
\toprule
Arm & Acc@0.5 & Acc@0.75 & Acc@0.9 & mAcc & Mean IoU \\
\midrule
$B$ (matched val baseline) & 89.218 & 81.029 & 62.753 & 76.475 & 82.557 \\
$R_{\mathrm{ALL}}$ & 90.328 & 83.581 & 69.394 & 79.514 & 84.327 \\
$CR_{\mathrm{ALL}}$ & 90.464 & 83.513 & 68.065 & 79.275 & 84.233 \\
\bottomrule
\end{tabular}%
}
\caption{Major~H RefCOCO validation-split unconditional-control readout
(26,488 rows; 2,738 images), scored on one matched row set with
10,000-iteration image-cluster bootstrap deltas against $B$.
$R_{\mathrm{ALL}}$ gains $+1.102$/$+2.552$/$+6.641$/$+3.039$ points at
Acc@0.5/0.75/0.9/mAcc (95\% CIs
$[+0.686,+1.519]$/$[+1.913,+3.181]$/$[+5.706,+7.574]$/$[+2.574,+3.508]$);
$CR_{\mathrm{ALL}}$ gains $+1.246$/$+2.484$/$+5.312$/$+2.800$ (95\% CIs
$[+0.820,+1.656]$/$[+1.849,+3.136]$/$[+4.394,+6.240]$/$[+2.347,+3.261]$).}
\label{tab:app-major-h-refcoco-val}
\end{table}

The specialist audit reuses the same row prefixes but records one timed repeat
after a per-row warm-up because EGM generation is substantially slower.  The
128- and 512-row gates both pass for all four arms: every expected key is
present, every timing is finite, and there is no OOM or unclassified runtime
failure.

\begin{table}[t]
\centering
\small
\setlength{\tabcolsep}{3.5pt}
\begin{tabular}{llrrrr}
\toprule
Rows & System & Calls & Gen. tokens & Mean s & P95 s \\
\midrule
128 & EGM-4B & 1.000 & 98.5 & 4.798 & 6.055 \\
128 & \quad + LFPR & 2.000 & 197.4 & 9.250 & 11.395 \\
128 & EGM-8B & 1.000 & 98.7 & 4.553 & 5.148 \\
128 & \quad + LFPR & 2.000 & 196.9 & 9.116 & 11.536 \\
\midrule
512 & EGM-4B & 1.000 & 98.9 & 4.622 & 5.503 \\
512 & \quad + LFPR & 1.998 & 197.9 & 9.080 & 10.861 \\
512 & EGM-8B & 1.000 & 97.8 & 4.581 & 5.764 \\
512 & \quad + LFPR & 2.000 & 196.5 & 9.154 & 10.531 \\
\bottomrule
\end{tabular}
\caption{Specialist cost stability across audit sizes.  Generated tokens are
summed across calls; each row has one synchronized timed repeat after a
per-row warm-up.}
\label{tab:app-egm-cost-stability}
\end{table}

At 512 rows, EGM-4B throughput changes from 0.216 to 0.110 rows/s and
maximum allocated memory remains 8.45~GiB; EGM-8B changes from 0.218 to 0.109
rows/s and remains at 16.48~GiB.  EGM-4B has one invalid row in both baseline
and LFPR (0.20\%); EGM-8B has none.  The paired mean overhead of LFPR is
4.458~s and 99.0 generated tokens for EGM-4B, and 4.574~s and 98.7 tokens for
EGM-8B.

A separate same-host anchor reruns the generalist $B$ and corrected $CRG$
arms on GPU~0 of the RTX~3090 host over the exact 128-row prefix.  It retains
the generalist protocol of one warm-up plus three timed repeats, giving 384
timed observations per arm.  $B$ has mean/p95 latency 1.041/1.185~s,
0.961 rows/s, 1.000 calls, 17.5 generated tokens, and 16.49~GiB peak allocated
memory.  $CRG$ has 1.948/2.284~s, 0.513 rows/s, 2.008 calls, 35.3 tokens, and
17.58~GiB.  Both arms have zero invalid outputs and zero runtime errors.  Only
one of the 128 rows is routed to the high-resolution pass, so this anchor is a
same-hardware cross-system comparison, not a substitute for the corrected
full-size generalist cost grid.  On these rows, EGM-4B/8B baseline latency is
$4.61\times$/$4.37\times$ $B$, and EGM-4B/8B with LFPR is
$4.75\times$/$4.68\times$ generalist $CRG$; generated-token ratios are about
$5.6\times$ in both comparisons.

The retrospective full-split effects by source are:

\begin{table}[t]
\centering
\small
\begin{tabular}{lrrr}
\toprule
Source & Rows & Acc@0.5 $\Delta$ & One-sided LCB \\
\midrule
COCO & 19,889 & +0.875 & +0.693 \\
Objects365 & 12,032 & +1.720 & +1.260 \\
\bottomrule
\end{tabular}
\caption{Full-split source-stratified effects.  Source rows use the same
frozen parser and paired image bootstrap; no interaction interval was archived.}
\label{tab:app-source}
\end{table}

The strict-IoU intervals for the full-split comparison are +3.173 points at
IoU~0.75 (LCB +2.831), +5.354 points at IoU~0.9 (LCB +4.898), and +3.066
points for mAcc (LCB +2.857).  They are exploratory because endpoint and
pipeline selection are not reflected in those historical one-sided bounds.

\subsection{Router diagnostics}

The label-free router flagged 463 of 2,000 pilot rows (23.2\%) and 6,641 of
31,921 full-split rows (20.8\%).  Where annotation boxes are available for
diagnosis, predicted-box buckets agree with annotation-defined buckets on
88.85\% of pilot rows and 89.83\% of full-split rows.  This statistic is not a
classifier accuracy and is not used as a gate; it describes why the
deployment proxy is a reasonable but imperfect substitute for the
annotation-stratified diagnostic.

\subsection{Malformed-row handling}

The pilot has zero runtime and invalid-prediction rows.  The full-split
baseline contains one inverted-coordinate completion.  The LFPR fusion
implementation passes an invalid incumbent through unchanged rather than
raising an exception or inventing a crop.  The resulting full-split report
therefore contains one inherited runtime/invalid count.  This row is
included in both matched arms and is not hidden from the error accounting.
It is a protocol repair for a degenerate baseline case, not a model or
hyperparameter change.

\subsection{Size-bucket baseline accuracy}
\label{app:size-baseline}

Table~\ref{tab:app-size-baseline} gives the raw, pre-intervention accuracy
by annotation-derived target size on the full 31,921-row test split; this
is the diagnostic that originally motivated targeting the small bucket
with additional resolution (Section~\ref{sec:diagnosis} below).

\begin{table}[t]
\centering
\small
\begin{tabular}{lrrr}
\toprule
Bucket & Rows (\%) & Acc@0.5 & Gap to large \\
\midrule
Small & 6,665 (20.9\%) & 78.92 & $-13.81$ \\
Medium & 16,311 (51.1\%) & 90.15 & $-2.58$ \\
Large & 8,945 (28.0\%) & 92.73 & -- \\
\bottomrule
\end{tabular}
\caption{Baseline Acc@0.5 by annotation-derived target size, before any
resolution intervention.}
\label{tab:app-size-baseline}
\end{table}

\subsection{Resolution-ladder and prompt-sensitivity pilot designs}
\label{app:ladder-designs}

All pilots in this subsection use 800 rows and the historical paired
image-level bootstrap.  The complete access chronology is not auditable, so
these comparisons are exploratory regardless of when individual runs
were launched.  The staged floor-selection summaries
(Section~\ref{sec:floor-selection} below) report a moderate floor of
$1{,}048{,}576$ pixels at $+0.75$ points relative to default (one-sided LCB
$-0.99$) and the chosen $4{,}194{,}304$ floor at $+1.875$ points relative to
default (LCB $+0.25$).  The
medium-bucket extension pilot applied the identical $4{,}194{,}304$-pixel
policy to an 800-row medium-bucket sample and gave $-0.125$ points (95\%
CI $[-1.86,+1.62]$).

The later prompt pilots (Section~\ref{sec:headroom-exhausted} below) used
an 800-row Objects365 sample paired against the same rows' recombined
post-floor predictions.  Two variants replaced the default prompt text without
changing any other setting: a \emph{tight-box} variant explicitly asking
for the tightest enclosing box gave $+0.125$ points (95\% CI
$[-1.39,+1.65]$, LCB $-1.13$), and a \emph{fine-grained} variant emphasizing
instance-level detail gave $+0.25$ points (95\% CI $[-1.25,+1.75]$, LCB
$-1.00$).  A further resolution rung at $8{,}388{,}608$ pixels
($\sim2\times$ the chosen floor), on the same 800-row small-bucket sample
used for the floor-selection ladder, gave 82.875\% versus 81.750\% at
$4{,}194{,}304$: a $+1.125$-point stepwise difference (95\% CI
$[-0.38,+2.73]$, LCB $-0.13$).  This value is not default-relative and
therefore cannot establish that the ladder peaked.  The former Figure~3 was
removed; row-level common-reference rescoring is required before drawing a
resolution-curve conclusion.

The zero-shot self-correction control
(Section~\ref{sec:self-correction-baseline} below) used the
same frozen incumbent with no fine-tuning: the model is shown the original
image with its own first-pass box rendered as an overlay and asked to
return one replacement box, with neither a guard nor a fusion step applied
to the output.  On its own 800-row pilot this gives Acc@0.5 $82.50\to80.50$
($-2.00$ points, LCB $-3.14$), Acc@0.75 $72.625\to70.50$ ($-2.125$ points,
LCB $-3.78$), Acc@0.9 $52.375\to54.625$ ($+2.25$ points, LCB $+0.12$), and
mAcc $68.35\to67.29$ ($-1.06$ points, LCB $-2.14$).

\subsection{Backbone-scaling pilot design}
\label{app:p0-design}

The backbone-scaling control in Table~\ref{tab:app-controls} uses the
official bf16 checkpoint of the same model family at approximately
$4\times$ the incumbent's parameter count, under the unchanged prompt,
parser, resolution policy, and evaluator.  A 500-row engineering pilot
completed with zero errors and clean bare-box parsing, confirming the
larger checkpoint is compatible with the frozen harness before spending a
larger budget.  The confirmation pilot used 380 rows (190 COCO, 190
Objects365, 344 unique images) rather than a larger round number: by this
point in the project, the shared image-disjoint validation pool had
already been drawn on by earlier pilots and related controls, and 380 was
the largest balanced size the remaining
unvisited pool supported without compromising image-disjointness from
every prior pilot.  On this confirmation pilot, Acc@0.9 regresses by
$-1.32$ points (LCB $-4.56$), the Objects365 subset effect is $-0.53$
points (failing outright), and the COCO LCB is $-2.07$ points, outside the
historical $-0.5$-point non-inferiority band.  No full-test run
was taken, and the FP8-quantized variant of the same checkpoint was not
separately piloted, since quantization would only be expected to add noise
in the same failing direction already observed at full precision.

The pilot above only tested the larger direction. To test the smaller
direction of the same model family -- a stronger test in one respect,
since a smaller same-architecture checkpoint reuses the incumbent's
frozen crop/guard/fusion policy directly rather than requiring new
adapter engineering -- we ran the complete $B/R/C/CR/CRG$ pipeline on
Qwen3-VL-4B-Instruct (roughly $0.5\times$ the incumbent's parameter
count, the identical \texttt{Qwen3VLForConditionalGeneration} architecture
class), under the unchanged prompt, parser, resolution floor, crop
context factor, and guard thresholds. It was evaluated on the same
3,000-row stratified Flickr30K merged-box subset used for the
cross-family screen below, with zero runtime errors on both the
direct-answer arm and the full guarded-fusion arm.
Table~\ref{tab:app-4b-scale} reports the paired result: every metric
improves, and every 95\% CI excludes zero, matching the direction (though
not the magnitude) of the 8B incumbent's own LFPR gain -- the mechanism
is not specific to the incumbent's particular parameter count.

\begin{table}[t]
\centering
\small
\resizebox{\linewidth}{!}{%
\begin{tabular}{lrrrr}
\toprule
Arm & Acc@0.5 & Acc@0.75 & Acc@0.9 & mAcc$_{0.50:0.95}$ \\
\midrule
$B$ (direct answer) & 72.933 & 57.500 & 36.633 & 54.703 \\
$+$LFPR ($CRG$) & 74.233 & 60.800 & 37.800 & 56.673 \\
$\Delta$ (95\% CI) & $+1.300$ $[0.494,\,2.122]$ & $+3.300$ $[2.297,\,4.349]$
& $+1.167$ $[0.236,\,2.085]$ & $+1.970$ $[1.377,\,2.572]$ \\
\bottomrule
\end{tabular}%
}
\caption{Qwen3-VL-4B-Instruct, full guarded-fusion pipeline vs.\ its own
direct-answer baseline, on the same 3,000-row stratified Flickr30K
merged-box subset used for the cross-family screen. All four 95\% paired
image-cluster bootstrap CIs exclude zero.}
\label{tab:app-4b-scale}
\end{table}

\subsection{Cross-family checkpoint screening}
\label{app:cross-family}

The backbone-scaling control above stays inside the incumbent's own model
family. To screen genuinely different architectures, we evaluated two
candidates' own direct-answer (arm $B$) accuracy before committing to the
substantially larger engineering effort of wiring a new checkpoint into the
full router/crop/guard/fusion pipeline, following the same B-arm-first
protocol already used elsewhere in this project to screen candidate
checkpoints cheaply: a candidate whose own accuracy sits meaningfully below
the incumbent offers nothing for LFPR to refine into a win.

\textbf{GLM-4.1V-9B-Thinking} (glm4v architecture, 9B) was evaluated on a
4,000-row proportionally stratified subset of the Flickr30K merged-box
confirmation (stratified on referring-expression length and ground-truth
box area, disjoint from no other evaluation in this paper). Table~\ref{tab:app-crossfamily}
reports its own Acc@0.5/0.75/0.9 and mAcc with an image-cluster bootstrap
CI, against the incumbent's own Acc@0.5 on the full 14,481-row population as
context (an unpaired reference, not a matched comparison). Its Acc@0.5 sits
$-4.69$ points below the incumbent's, with the 95\% CI excluding the
incumbent's point estimate. Two further practical problems compound this:
box-parser success is only $83.1\%$ ($3{,}323/4{,}000$ rows), well below the
incumbent's own $99.6\%$ (Table~\ref{tab:app-cost-r3}'s invalid-output
rate), and per-row latency has a severe unpredictable tail (up to $153.4$s
on a single row against a $9.2$s mean) driven by unbounded ``thinking''
trace length -- a cost profile that would compound badly once the
guard/crop pipeline's extra forward passes are layered on top.

\textbf{Qwen2.5-VL-7B-Instruct} (a different generation and architecture
within the Qwen family) was evaluated earlier in this project as an
independent harness-reconciliation anchor: on an identical 3,000-row random
subsample of Ref-L4-test, it reaches Acc@0.5 $=81.57\%$
($2{,}910/3{,}000$ valid predictions, $97.0\%$ parser success) against the
incumbent's own Acc@0.5 $=88.03\%$ on the same rows through the same
harness -- a $-6.47$-point paired, same-harness gap. Its own reproduction
also lands closer to its publisher's reported Ref-L4-test number ($+0.77$
points offset) than GLM-4.1V-9B-Thinking's reproduction does ($-2.90$
points offset), indicating a more faithfully reproduced evaluation.

\begin{table}[t]
\centering
\small
\begin{tabular}{lrrrrr}
\toprule
Checkpoint & Dataset (rows) & Acc@0.5 & Parser success & Incumbent gap \\
\midrule
GLM-4.1V-9B-Thinking & Flickr merged-box (4,000) & 70.18 & 83.1\% & $-4.69$pp \\
Qwen2.5-VL-7B-Instruct & Ref-L4-test (3,000, paired) & 81.57 & 97.0\% & $-6.47$pp \\
\bottomrule
\end{tabular}
\caption{Cross-family checkpoint screening summary. The two rows use
different evaluation datasets (noted in column 2) and are not directly
comparable to each other in absolute accuracy; each is compared only to
the incumbent on its own dataset. Parser success is the fraction of rows
producing a usable bare box.}
\label{tab:app-crossfamily}
\end{table}

Neither candidate's own accuracy sits close enough to the incumbent's to
justify building the full LFPR adapter under this project's own established
screening rule, so no crop/guard/fusion pipeline was built for either
checkpoint. Between the two, Qwen2.5-VL-7B-Instruct is the more
procedurally viable candidate -- higher parser reliability, no latency
pathology, and closer harness fidelity to its own published number -- and
would be the natural starting point if a future revision extends this
screening with a full second-checkpoint LFPR run. We report both attempts
in full rather than omitting the negative result, consistent with this
paper's evidence-tier discipline of not repackaging a screening pilot as
confirmatory evidence.

\subsection{Why no head-to-head run against recent training-free baselines}
\label{app:baseline-incompatibility}

Section~\ref{sec:related} notes that Chain-of-Caption is complementary to
our geometry-based accept/reject step rather than a substitute for it.  We
additionally checked compatibility for a head-to-head run: Chain-of-Caption's
paper reports results on NVIDIA VILA and Qwen2.5 backbones, not Qwen3-VL,
and we were unable to locate a public implementation compatible with our
Qwen3-VL pipeline through the paper's links, an author repository, or
standard search; absent a repository URL from the authors, we do not claim
that no such implementation exists, only that we did not find one, so no
head-to-head run was attempted.
The same question arises for the two other closest training-free methods
on the same benchmark family, Entropy-Gradient Grounding (EGG) and
Test-time Scaling over Perception (TTSP); this appendix expands the
compatibility check behind that decision rather than leaving it asserted.

\textbf{EGG} \citep{gropl2026entropygradient} back-propagates
next-token entropy to visual token embeddings to produce a grounding
signal without an auxiliary detector. Its public released code supports
inference for LLaVA~1.5/1.6 only; the authors' own repository lists
Qwen~2.5 and InternVL~3.5 support as not-yet-released roadmap items, and
this project's incumbent (Qwen3-VL-8B-Instruct) is not on that roadmap at
all. Running EGG would therefore mean either evaluating it on a
completely different backbone (confounding the comparison with a model
change, not just a method change) or reimplementing its entropy-gradient
mechanism for Qwen3-VL from the paper alone, which we judged too close to
guessing the authors' intended implementation to report as a faithful
reproduction. The repository also carries no stated license, an
independent reason not to build on it without contacting the authors.

\textbf{TTSP} \citep{jiang2026testtimeperception} generates multiple
perceptual exploration traces, filters them by entropy-based confidence,
and iteratively refines a structured evidence ledger. Its public
Apache-2.0-licensed code does support Qwen3-VL-8B-Instruct directly --
architecturally, this is the more compatible of the two candidates. The
incompatibility is instead at the task-interface level: TTSP's released
pipeline and benchmark CLI take a closed-set multiple-choice question
(\texttt{question} plus an \texttt{options} list) and return a
reliability-weighted vote over those options; it has no open bounding-box
output mode. Adapting it to referring-expression grounding would mean
replacing its core voting mechanism (which tallies votes over discrete
options) with a continuous box-aggregation rule of our own design -- at
that point the result would be a new method inspired by TTSP's
evidence-ledger idea, not a reproduction of TTSP itself, and reporting it
as ``TTSP's accuracy on this benchmark'' would misattribute our own design
choices to the original authors.

Both checks were run against the actual released code and documentation,
not assumed from the papers alone, before concluding neither supports a
faithful head-to-head comparison on this task without either changing the
backbone (EGG) or replacing the core method (TTSP).

\section{Control Experiments and Stopping Rules}
\label{app:controls}

The following table records the nearby routes that were evaluated on
image-disjoint pilots or reserved evaluation data.  ``Closed'' records the
historical stopping decision under the then-used one-sided gate; it does not
denote a multiplicity-adjusted confirmatory conclusion.

{\small
\begin{longtable}{p{0.24\linewidth}p{0.19\linewidth}p{0.45\linewidth}}
\caption{Matched controls and dispositions.}
\label{tab:app-controls} \\
\toprule
Route & Main result & Disposition \\
\midrule
\endfirsthead
\multicolumn{3}{l}{\itshape Table~\ref{tab:app-controls} continued from previous page} \\
\toprule
Route & Main result & Disposition \\
\midrule
\endhead
\midrule
\multicolumn{3}{r}{\itshape continued on next page} \\
\endfoot
\bottomrule
\endlastfoot
Released EGM transfer & $-7.45$ points at Acc@0.5;
95\% CI $[-9.56,-5.34]$ & Closed as a zero-shot transfer; native EGM recipe still needs a matched-compute replication. \\
Supervised coarse-to-fine & $-0.543$ points at Acc@0.5;
LCB $-1.264$ & Closed before the planned larger run. \\
Transferred parallel box decoding & $-15.80$ points;
LCB $-17.64$ & Transfer route closed. \\
SAM3 projection & Positive strict-IoU effects but no headline gate & Closed as a standalone replacement. \\
SAM3 midpoint fusion & $+0.10$ points at Acc@0.5;
LCB $-0.20$ on fresh data & Retrospective fusion signal did not replicate. \\
Full crop replacement & $+0.50$ points at Acc@0.5;
LCB $-0.10$ & Crop replacement closed; strict-IoU signal retained. \\
Standalone midpoint fusion & $+0.70$ points on the reused pilot;
source-weighted projection $+0.65$ & Useful component; insufficient alone. \\
LFPR & $+1.194$ points on full test;
LCB $+0.984$ & Strongest matched full-split result in the audit; retrospective. \\
Frozen-split RefCOCO-family transfer & $+0.817$ points on pooled RefCOCO/+/g;
95\% CI $[+0.639,+1.001]$ & Confirmed positive transfer at Acc@0.5; Acc@0.9 is
 unchanged, not because an unguarded arm banks the strict-IoU gain instead
 (the genuine unguarded negative control underperforms the incumbent on
 every metric, Table~\ref{tab:app-e3-arms}), but because the guard trades a
 small Acc@0.9 concession for the Acc@0.5 gain directly; a strict
 unrelated-model-family transfer control was cancelled as redundant with this
 result, and the released-specialist baseline comparison is reported
 separately in Section~\ref{sec:specialists}. \\
Backbone scaling (32B) & $+0.26$ points at Acc@0.5;
LCB $-2.45$ & Scaling closed; fails on both data sources and regresses Acc@0.9. \\
Backbone scaling (4B, full LFPR sweep) & $+1.970$ points mAcc$_{0.50:0.95}$;
95\% CI $[+1.377,+2.572]$ & Confirmed positive; the mechanism generalizes
to a $4\times$-smaller same-family checkpoint (Appendix~\ref{app:p0-design}). \\
Guard/fusion threshold sweep & No configuration Pareto-dominates the shipped
rule across a max-zoom-consistency grid of $0.15$--$1.00$ and a fusion-weight
grid of $0.5$--$1.0$ & Closed: every accuracy-improving configuration fails a
Holm-adjusted significance check; the shipped threshold is not improvable
within the swept grid. \\
Confidence-gate mechanism control (Ref-L4, $R$ incumbent) & An independent
crop-refinement pass over all 31{,}921 Ref-L4 $R$ rows, recording both the
shipped geometry guard's admission decision and the model's own
greedy-decoding confidence (mean token log-probability) on every row, then
scoring the same replacement/midpoint-fusion formulas gated by each rule in
turn (image-clustered, 9{,}467 images).  Geometry-gated fusion reproduces a
small Acc@0.5 gain over $R$, $[+0.139,+0.399]$ (95\% CI), with the same
Acc@0.9/mAcc cost already reported for $CRG-R$ on the RefCOCO family; the
confidence-gated analogue does not reproduce even that Acc@0.5 gain,
$[-0.334,-0.098]$, and geometry-gated replacement-only is far worse than
either fusion arm, mAcc $[-2.987,-2.437]$.  The two admission rules agree on
$52.5\%$ of rows ($15{,}411$ both-admit, $1{,}355$ both-reject) &
Closed: on this independent replication, a frozen label-free confidence
threshold is not a substitute for the shipped geometry guard -- only the
geometry-gated arm recovers the small Acc@0.5 gain over routing, and both
share the Acc@0.9/mAcc trade-off already attributed to the post-routing
marginal effect (Appendix~\ref{app:e3-transfer}; \S\ref{sec:flickr30k}), so
agreement on \emph{which} rows to admit does not by itself explain the
guard's contribution. \\
Cross-family incumbent candidate (zero-shot, pre-refinement) & Full 31,921-row
result: Acc@0.5 $53.933$, Acc@0.75 $39.958$, Acc@0.9 $22.22$, and mean IoU
$49.641$, versus $88.531$ Acc@0.5 for the Qwen3-VL-8B incumbent; 6,584 rows
($20.6\%$) were unparseable or malformed and scored as zero IoU.  The full
split confirms the 500-row pilot's negative direction (pilot Acc@0.5 $50.60$). &
Closed at the zero-shot stage; the pilot and full-split directions agree, so
the candidate was not carried into the refinement pipeline. \\
Full-parameter fine-tune & Image-disjoint 2,000-row held-out evaluation:
Acc@0.5 $87.50\to86.05$ ($-1.45$), Acc@0.75 $76.15\to74.55$ ($-1.60$),
Acc@0.9 $54.70\to54.40$ ($-0.30$) & Closed: underperforms the frozen
incumbent at every threshold; the training loss plateaus after warmup,
consistent with the negative held-out result. \\
Iterative zoom (2 unconditional rounds, no guard) & Acc@0.5 $89.00\to86.80$
($-2.20$), Acc@0.9 $55.60\to51.20$ ($-4.40$) on a 500-row sample & Closed:
repeated unguarded cropping compounds error rather than refining it. \\
Random- and center-crop placement controls (matched call count) & Acc@0.5
collapses to $42.00$ (random) and $44.20$ (center) vs.\ $89.00$ incumbent on
the same 500-row sample & Supports the mechanism claim: the crop's gain
requires an incumbent-informed window, not merely a second look. \\
Self-consistency, matched call count (2 samples, temperature $0.7$,
midpoint-fused) & Acc@0.5 $89.00\to88.40$, Acc@0.9 $55.60\to52.60$ on the
same 500-row sample & Supports the mechanism claim: re-sampling the
identical question does not recover the crop's gain. \\
\end{longtable}
}

The controls are not presented as a universal impossibility theorem.  They
close the tested instantiations under the declared protocol and justify
keeping the main claim narrow.  A substantially larger backbone within the
same model family (Qwen3-VL-32B, roughly $4\times$ the parameters) was also
evaluated under the identical frozen prompt, parser, and evaluator; it
failed every admission requirement, with a positive but non-significant
Acc@0.5 point estimate (LCB negative), a regression at the stricter
Acc@0.9 threshold, and a failing effect on the Objects365 subset
(Table~\ref{tab:app-controls}).  Scaling the backbone within this family
therefore does not reliably close the residual gap under this harness.
LFPR's positive result does not imply that all spatial heads or proposal
sources would fail; it shows that this specific label-free composition is
useful without reaching absolute SOTA.

\begin{figure}[t]
\centering
\includegraphics[width=\linewidth]{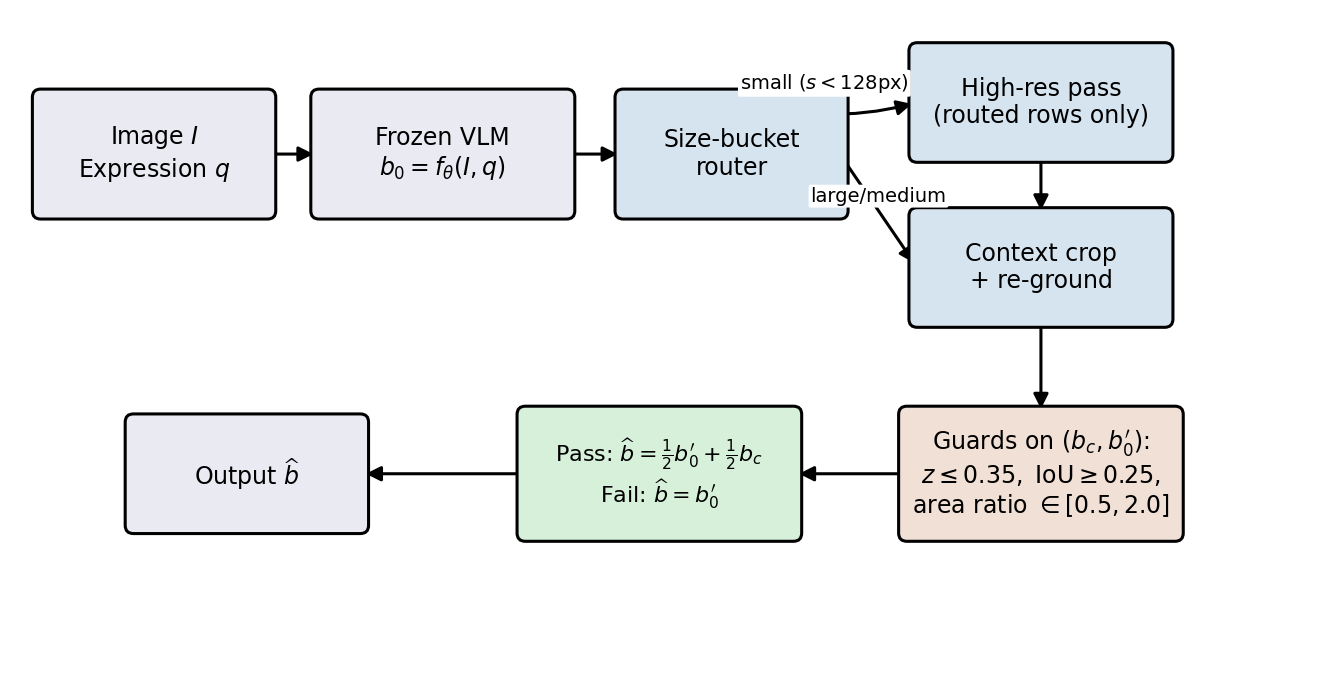}
\caption{The LFPR pipeline.  The router decision and the guard checks are
both computed from already-observed model outputs and image geometry, never
from the ground-truth box.}
\label{fig:pipeline}
\end{figure}

\section{Procedure in pseudocode}
\label{app:algorithm}

The prose in Section~\ref{sec:method} is normative; this restatement is
provided for implementers.

\begin{algorithm}[t]
\caption{Label-free precision refinement}
\label{alg:lfpr}
\begin{algorithmic}[1]
\STATE \textbf{Input:} image $I$, expression $q$, frozen VLM $f_\theta$
\STATE $b_0 \leftarrow \operatorname{canon}(f_\theta(I,q))$
       \COMMENT{parse, clip, order, validate}
\IF{$b_0$ is invalid}
  \STATE \textbf{return} $b_0$ \COMMENT{retained as a miss, never synthesized}
\ENDIF
\STATE $s \leftarrow \sqrt{\max(0,x_2{-}x_1)\max(0,y_2{-}y_1)}$ for $b_0$
\STATE \COMMENT{label-free size-bucket router, decided from $b_0$ alone}
\IF{$s < 128$ px}
  \STATE $b_r \leftarrow \operatorname{canon}(f_\theta(I,q \mid
         \text{min-pixel budget }4{,}194{,}304))$
  \STATE $b_0' \leftarrow b_r$ if valid, otherwise $b_0$
\ELSE
  \STATE $b_0' \leftarrow b_0$
\ENDIF
\STATE $C \leftarrow$ integer-clipped context crop of $b_0'$ with $\gamma{=}2.0$
\STATE $b_c \leftarrow \operatorname{canon}($map $f_\theta(C,q)$ to image$)$
\IF{$b_c$ is valid and $z(b_c,b_0')\le0.35$ and
    $\iou(b_c,b_0')\ge0.25$ and $A(b_c)/A(b_0')\in[0.5,2.0]$}
  \STATE $\widehat b \leftarrow \operatorname{canon}(\tfrac{1}{2}b_0'
         + \tfrac{1}{2}b_c)$
  \IF{$\widehat b$ is invalid}
    \STATE $\widehat b \leftarrow b_0'$
  \ENDIF
\ELSE
  \STATE $\widehat b \leftarrow b_0'$ \COMMENT{guard failure: keep the pre-crop answer}
\ENDIF
\STATE log route, parse outcomes, guard values, and final decision
\STATE \textbf{return} $\widehat b$
\end{algorithmic}
\end{algorithm}

\begin{table*}[t]
\centering
\small
\begin{tabular}{lrrrr}
\toprule
Metric & Incumbent & LFPR & $\Delta$ (pp) & Two-sided 95\% CI \\
\midrule
Acc@0.5 & 88.531 & 89.725 & +1.194 & [0.944, 1.439] \\
Acc@0.75 & 77.416 & 80.590 & +3.173 & [2.768, 3.584] \\
Acc@0.9 & 55.788 & 61.142 & +5.354 & [4.808, 5.906] \\
mAcc@0.5:0.95 & 72.947 & 76.013 & +3.066 & [2.818, 3.311] \\
Mean IoU & 80.944 & 82.648 & +1.704 & [1.542, 1.868] \\
\bottomrule
\end{tabular}
\caption{Matched full-split results on 31,921 expressions from 9,467 images.
Intervals use 10,000 image-cluster bootstrap resamples and describe this
retrospective comparison.  $\macc_{0.5:0.95}$ is the manuscript's primary
summary, designated retrospectively rather than confirmed by an external
preregistration; Acc@0.5 is the standard benchmark operating point;
Acc@0.75/0.9 are secondary diagnostics; Mean IoU is reported alongside as
a threshold-free summary.  All five are exploratory and unadjusted here
because the comparison itself is retrospective, not because of their role
in the metric hierarchy.}
\label{tab:main-results}
\end{table*}

\subsection{Routing-alone and post-routing marginal decomposition}
\label{app:refl4-decomposition}

RefCOCO-family (Table~\ref{tab:app-components}) and the Flickr30K
merged-box confirmation (Table~\ref{tab:app-flickr-mechanism}) both report
a same-row decomposition that isolates label-free resolution routing's own
effect ($R$-$B$) from the marginal effect of crop, guard, and fusion on top
of it ($CRG$-$R$). Table~\ref{tab:app-refl4-decomposition} reports the
identical decomposition for Ref-L4, computed with zero new inference from
the same three already-archived, hash-verified per-row artifacts behind
Table~\ref{tab:main-results} (\texttt{audit\_refl4\_contrasts.py}; the
script cross-checks the fused artifact's embedded $R$ box against the
standalone $R$ artifact's own prediction on every row before scoring, and
refuses to report a number unless the recomputed $B$/$CRG$ Acc@0.5
reproduce Table~\ref{tab:main-results} to within $0.01$pp -- both gates
passed here).

\begin{table}[t]
\centering
\small
\begin{tabular}{lrrrr}
\toprule
Contrast & Acc@0.5 & Acc@0.75 & Acc@0.9 & mAcc \\
\midrule
$R$-$B$ (routing alone) & $+0.711$ & $+1.920$ & $+3.255$ & $+1.773$ \\
& $[0.498, 0.932]$ & $[1.607, 2.244]$ & $[2.862, 3.658]$ & $[1.560, 1.994]$ \\
$CRG$-$R$ (marginal, after routing) & $+0.482$ & $+1.253$ & $+2.099$ & $+1.293$ \\
& $[0.347, 0.626]$ & $[0.974, 1.535]$ & $[1.669, 2.540]$ & $[1.154, 1.431]$ \\
$CRG$-$B$ (deployed policy) & $+1.194$ & $+3.173$ & $+5.354$ & $+3.066$ \\
& $[0.944, 1.439]$ & $[2.768, 3.584]$ & $[4.808, 5.906]$ & $[2.818, 3.311]$ \\
\bottomrule
\end{tabular}
\caption{Ref-L4's routing-alone and post-routing marginal contrasts,
31{,}921 rows, 9{,}467 images, 10{,}000-iteration image-cluster bootstrap
(95\% CI below each point estimate). $CRG$-$B$ reproduces
Table~\ref{tab:main-results} exactly, as it must since both are computed
from the same artifacts. Unlike the RefCOCO family (where $CRG$-$R$'s
Acc@0.9 is negative, $-1.192$pp, and unlike Flickr30K (where
crop/guard/fusion adds no accuracy beyond routing at any threshold,
\S\ref{sec:flickr30k}), Ref-L4 is the one evidence tier where
crop/guard/fusion has a genuine positive marginal effect beyond routing at
every reported threshold, not merely a small Acc@0.5-for-Acc@0.9 trade:
$CRG$-$R$'s Acc@0.9 is $+2.099$pp, not negative. Routing alone still
accounts for the majority of the deployed policy's Acc@0.9 effect
($+3.255$ of $+5.354$pp, 61\%), so the same qualitative pattern found on
the other two tiers -- routing does most of the strict-IoU work -- still
holds directionally here, but Ref-L4 is not simply a case of the two
stages offsetting each other.}
\label{tab:app-refl4-decomposition}
\end{table}

\subsection{Guard-by-fusion factorial decomposition}
\label{app:m2-factorial}

$CRG$ crosses two binary choices: whether the guard gates the crop
candidate at all, and whether an admitted candidate replaces the incumbent
outright or is midpoint-fused with it. Because the incumbent box, the crop
candidate, and the guard's own admit/reject decision are already stored
per row, all four cells of this $2\times2$ design are computable offline
from the same frozen artifact with no additional inference, at the exact
input denominator (31,921 rows, 1 invalid incumbent, 29,523 rows with an
admitted candidate).

\begin{table*}[t]
\centering
\small
\resizebox{\linewidth}{!}{%
\begin{tabular}{lrrrrrl}
\toprule
Cell & Acc@0.5 & Acc@0.75 & Acc@0.9 & mAcc & Mean IoU $\Delta$ & IoU W/L/T \\
\midrule
Guarded, midpoint ($CRG$, shipped) & \textbf{89.725} & 80.590 & 61.142 & 76.013 & +1.704 & 19{,}386 / 10{,}260 / 2{,}275 \\
Guarded, replacement & 89.399 & 79.947 & \textbf{63.143} & 76.166 & +1.725 & 17{,}333 / 12{,}317 / 2{,}271 \\
Unguarded, midpoint & 88.262 & 77.739 & 59.350 & 73.864 & +0.559 & 19{,}764 / 11{,}321 / 836 \\
Unguarded, replacement & 87.247 & 78.140 & 62.059 & 74.491 & +0.129 & 17{,}612 / 13{,}490 / 819 \\
\bottomrule
\end{tabular}%
}
\caption{Guard $\times$ fusion factorial on the corrected full split
(Appendix~\ref{app:erratum}; 10,000-iteration image-cluster bootstrap per
cell). Mean IoU $\Delta$ and IoU W/L/T are against the true pre-resolution
incumbent $B$, cross-checked against the standalone $B$ artifact
(Appendix~\ref{app:refl4-meaniou-erratum}); an earlier version of this
table computed both columns against the post-routing $R$ box while this
caption described them as against $B$, which understated every cell's
Mean IoU gain and, for the two unguarded cells, had the wrong sign. The
guard, not the
fusion rule, is what protects Acc@0.5: both guarded cells beat both
unguarded cells at Acc@0.5, and only the unguarded cells fall below the
88.531\% incumbent. Midpoint fusion trades some of the guarded arm's
Acc@0.9 for stability -- guarded replacement alone reaches 63.143\% at
Acc@0.9 (its 95\% CI is $[+3.512,+4.700]$pp over the incumbent, wider than
$CRG$'s $[+1.662,+2.545]$pp) at the cost of a slightly lower Acc@0.5 and
more harmed rows (12,317 versus $CRG$'s 10,260). $CRG$ is the
historically fixed shipped policy -- its parameters were set before this
factorial was computed, not selected by it; the other three cells are
reported for mechanism decomposition, not as alternate deployment
candidates.}
\label{tab:app-m2-factorial}
\end{table*}

This factorial resolves a concern that the two-arm $CRG$-versus-guard-off
contrast alone cannot separate rejection from midpoint shrinkage as the
source of the accuracy trade: with
both binary factors held out, rejection (the guard) is the factor that
protects Acc@0.5, and midpoint fusion (independent of the guard) is what
gives up strict-IoU accuracy for stability rather than for its own
accuracy gain, since guarded replacement's Acc@0.9 exceeds guarded
midpoint's by 2.0 points at a comparable Acc@0.5.

\paragraph{Direct paired contrast between the two guarded cells.} The two
guarded-cell CIs above are each against the incumbent $B$, which does not
by itself establish whether guarded replacement and guarded midpoint
($CRG$) differ from each other.  We therefore computed the direct paired
contrast between exactly these two cells on the same 31,921 rows (10,000
image-cluster bootstrap resamples; zero new model inference, reusing the
same incumbent/candidate/guard fields already stored for the table above;
\texttt{recompute\_refl4\_replacement\_vs\_midpoint.py}, which refuses to
report a number unless the recomputed guarded-replacement and
guarded-midpoint scores both reproduce this table's published figures to
within $0.01$pp -- both gates passed here, matching to within $0.001$pp).
Guarded replacement minus $CRG$ is $-0.326$ points at Acc@0.5 (95\% CI
$[-0.447,-0.208]$), $-0.642$ at Acc@0.75 (CI $[-0.933,-0.343]$), $+2.002$
at Acc@0.9 (CI $[1.600,2.420]$), and $+0.153$ at mAcc (CI
$[0.011,0.299]$, $p=0.034$).  All four intervals exclude zero: guarded
replacement is significantly better than the shipped policy at Acc@0.9
and at mAcc -- the manuscript's own declared primary endpoint -- and
significantly worse at Acc@0.5/0.75.  $CRG$ remains the shipped policy
because it was the historically fixed choice before this factorial was
computed (\S\ref{sec:prereg}), not because it dominates guarded
replacement on every metric; it does not, and this direct contrast is
reported so that a reader who weights strict-IoU accuracy or the primary
endpoint above Acc@0.5 can see exactly where the two policies disagree
and by how much.

\subsection{IoU transition matrix}
\label{app:transition-matrix}

\begin{figure}[t]
\centering
\includegraphics[width=\linewidth]{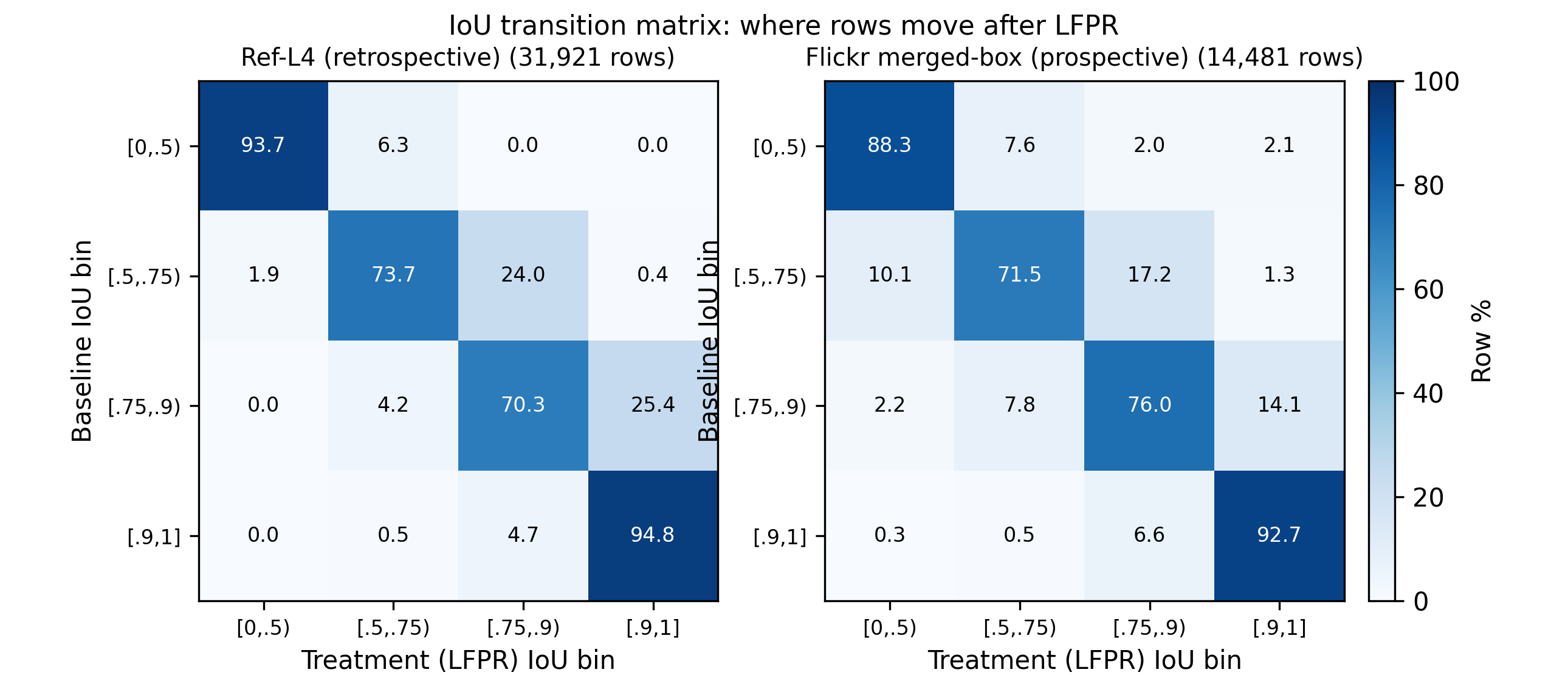}
\caption{IoU transition matrix. Diagonal cells dominate on both datasets
(70--95\% of rows stay in their baseline bin), consistent with LFPR
changing a minority of predictions (Section~\ref{sec:diagnosis}). The two
off-diagonal cells of most interest: strict-IoU harm, $[.9,1]\to[.75,.9)$,
affects $4.7\%$ of Ref-L4's near-perfect rows ($894/18{,}847$) and $6.6\%$
of Flickr's ($359/5{,}467$) -- a real, nontrivial cost, not a rounding
artifact, consistent with the guard-versus-Acc@0.9 trade reported
throughout this paper. Genuine boundary refinement, $[.5,.75)\to[.9,1]$,
is comparatively rare on both datasets ($0.4\%$ Ref-L4, $1.3\%$ Flickr);
most of LFPR's gain instead comes from the larger
$[.75,.9)\to[.9,1]$ cell ($25.4\%$ Ref-L4, $14.1\%$ Flickr), i.e.\ rows
that were already close and get pushed over the strict threshold, not
rows recovered from a coarse miss.}
\label{fig:transition-matrix}
\end{figure}

Figure~\ref{fig:transition-matrix} bins baseline (pre-LFPR) and treatment
(post-LFPR) IoU into $[0,.5)$, $[.5,.75)$, $[.75,.9)$, $[.9,1]$ and reports
the row-normalized transition matrix for Ref-L4 and the Flickr merged-box
confirmation, using the same paired per-row predictions underlying
Table~\ref{tab:app-e3-arms} and Table~\ref{tab:app-flickr-mechanism} --
no new inference.

\subsection{Qualitative cases}
\label{app:e5-qualitative}

Rows below are selected by a rule fixed before inspection: the three
largest per-row IoU gains and three largest losses among admitted-candidate
rows, plus the rates of the two out-of-policy paths a reader might expect
but that are not selected as favourable cases. Guard rejections (2,394 of
31,921 rows, 7.5\%) and parser failures (3 rows, 0.009\%) are reported by
pool size rather than by example, since neither category is about the
fused box's accuracy.

\begin{table*}[t]
\centering
\small
\resizebox{\linewidth}{!}{%
\begin{tabular}{lp{0.55\linewidth}rr}
\toprule
Category & Expression (truncated) & Incumbent IoU & Fused IoU \\
\midrule
Largest improvement & The middle white boat docked between two other vessels, distinguished\ldots & 0.561 & 0.891 \\
Largest improvement & The table, adorned with a white tablecloth, has a man in a gray shirt\ldots & 0.657 & 0.958 \\
Largest improvement & The table covered with a white tablecloth, at which the man wearing a\ldots & 0.644 & 0.933 \\
Largest harm & An individual in a blue and black jersey is maneuvering the foremost b\ldots & 0.985 & 0.574 \\
Largest harm & Within the cluttered area, a building instrument, namely a hammer, is\ldots & 0.940 & 0.530 \\
Largest harm & A cluster of thin green onions in a blue container. & 0.954 & 0.558 \\
\bottomrule
\end{tabular}%
}
\caption{Fixed-rule qualitative cases from the corrected full split
(Appendix~\ref{app:erratum}).
The largest-harm rows share a pattern: the incumbent was already close to
correct (IoU $>0.9$), and the crop candidate pulled the box toward a
plausible but wrong sub-region (a hand, a nearby object, a neighbour in a
cluster) that the guard's geometric checks did not catch. The
largest-improvement rows are the mirror case: a coarse incumbent box that
the crop resolves to the correct sub-region within a larger scene. Neither
pattern is unique to one expression style or object category in this
sample.}
\label{tab:app-e5-cases}
\end{table*}

\subsection{Erratum: the silently discarded pixel budget}
\label{app:erratum}

An earlier version of this work reported that the resolution router fired on
2,223 transfer rows without altering a single output, and attributed the null
to RefCOCO-family images already exceeding the routed pixel budget.  Both
parts were wrong.  The images are roughly fifteen times \emph{smaller} than
that budget, not larger, and the budget was never applied: the override was
passed as a processor keyword argument that this checkpoint's fast image
processor silently discards, leaving the checkpoint default in force.  The
routed pass therefore ran at default resolution and reproduced the incumbent
byte for byte, which is also why all five arms of the cost audit reported an
identical 16.49~GiB peak, i.e.\ model weights only.

Five call sites carried the defect.  All are repaired, and the budget is now
written into the resolved image-processor configuration and verified, so a
request that fails to take effect raises an error instead of silently
producing a no-op arm.  Every resolution-routed number in this paper was re-measured
afterwards.  On the transfer split the regenerated arm changes 2,213 of 2,223
flagged rows against zero before; on Ref-L4 it changes 6,591 of 6,641 against
627 rows of $\pm1$-unit numerical jitter before.

\subsection{Erratum: Ref-L4's Mean IoU was reported against $R$, not $B$}
\label{app:refl4-meaniou-erratum}

Two tables reported Ref-L4's Mean IoU delta as though it were the deployed
policy's effect against the pre-resolution incumbent $B$
(Table~\ref{tab:main-results}; Table~\ref{tab:evidence-tiers}'s Ref-L4 row),
and a third (Table~\ref{tab:app-m2-factorial}) reported both its Mean IoU
delta and win/loss/tie columns the same way for all four factorial cells.
In every case the reported ``incumbent'' Mean IoU was actually the
post-routing $R$ arm's Mean IoU (81.957\%): a fresh, hash-verified audit
against the standalone pre-resolution $B$ artifact
(\texttt{audit\_refl4\_contrasts.py}, run while adding the $R$-alone/$CRG$-$R$
decomposition in Table~\ref{tab:app-refl4-decomposition} below) finds $B$'s
true Mean IoU is 80.944\%, not 81.957\% -- the same order of mistake as
Blocker~1 in the review history, here isolated to the Mean IoU column of
three tables rather than a whole mislabeled file. Acc@0.5/0.75/0.9/mAcc were
never affected: those columns were always computed from the correct $B$
artifact, which is exactly how the discrepancy was caught -- the audit
script's identity gate reproduces every other Ref-L4 figure in
Table~\ref{tab:main-results} to within $0.001$pp, so the artifact is not in
question, only which arm's Mean IoU had been transcribed into the
``incumbent'' column. All three tables are corrected in this revision to
report $B$'s true 80.944\% Mean IoU and the resulting deltas
(Table~\ref{tab:main-results}: $+1.704$pp, not $+0.691$pp;
Table~\ref{tab:app-m2-factorial}: $+1.704$/$+1.725$/$+0.559$/$+0.129$pp for
the four cells, not $+0.691$/$+0.712$/$-0.454$/$-0.884$pp -- the two
unguarded cells' sign flips from apparently harming Mean IoU relative to
$B$ to slightly helping it, though both still trail the guarded cells by a
wide margin and neither changes the guard's role in protecting Acc@0.5).
The separate $80.936\%\to81.949\%$ routing-only figures in
\S\ref{sec:floor-selection} are a distinct, already-flagged superseded
diagnostic from before the pixel-budget fix above, not this erratum;
that section's canonical cross-reference is corrected to $80.944\%$
accordingly.

\section{Detail relocated from the main text}
\label{app:relocated}

The subsections below were condensed out of the main text, either in this
revision cycle or an earlier one, to meet the page limit.  They are
retained verbatim because each is cited from the main text, but none is
needed to follow the argument there: they record cost tables, protocol
detail, exploratory diagnostics, and closed negative controls.
\subsection{Specialist cost cross-check on RTX~3090}

The generalist cost audit reported in Appendix~\ref{app:cost-audit} runs
entirely on the same RTX~3090 host used throughout this paper; an earlier
draft additionally measured the generalist arms on a single RTX~PRO~6000
before the specialist checkpoints were available on that host, but that
measurement used different hardware from every other cost number in this
paper and has been withdrawn to avoid implying a same-hardware comparison
that was never intended.  The specialist cross-check below stays on
RTX~3090 throughout, matching Appendix~\ref{app:cost-audit}'s protocol
exactly.

A specialist audit used four physical RTX~3090 GPUs, with each arm
split into two non-overlapping shards and one worker per GPU.  EGM-4B and
EGM-8B were each evaluated alone and with LFPR using batch size one, bfloat16,
SDPA attention, greedy decoding, one per-row warm-up, and one synchronized
timed repeat.  All four arms passed the 128- and 512-row admission gates with
complete keys, finite timings, no OOM, and no unclassified runtime failure.

\begin{table*}[t]
\centering
\small
\setlength{\tabcolsep}{3.5pt}
\begin{tabular}{lrrrrrrr}
\toprule
System & Calls/row & Gen. tokens & Mean s & P95 s & Rows/s & Peak GiB & Invalid \\
\midrule
EGM-4B & 1.000 & 98.9 & 4.622 & 5.503 & 0.216 & 8.45 & 0.20\% \\
\quad + LFPR & 1.998 & 197.9 & 9.080 & 10.861 & 0.110 & 8.45 & 0.20\% \\
EGM-8B & 1.000 & 97.8 & 4.581 & 5.764 & 0.218 & 16.48 & 0.00\% \\
\quad + LFPR & 2.000 & 196.5 & 9.154 & 10.531 & 0.109 & 16.48 & 0.00\% \\
\bottomrule
\end{tabular}
\caption{Synchronized 512-row specialist cost audit on physical RTX~3090
GPUs.  Generated tokens are summed across calls.  Peak memory is maximum
allocated memory; invalid rows remain in the denominator.  Each row has one
timed repeat after a per-row warm-up.}
\label{tab:egm-cost}
\end{table*}

For EGM-4B, LFPR adds 99.0 generated tokens and 4.458~s per paired row,
giving $2.001\times$ the generated tokens and $1.965\times$ the latency of the
specialist alone.  For EGM-8B, the corresponding overhead is 98.7 tokens and
4.574~s, or $2.009\times$ the tokens and $1.999\times$ the latency.  Throughput
falls by approximately one half in both cases.  Sequential refinement does
not add model-resident memory: maximum allocated memory remains 8.45~GiB for
EGM-4B and 16.48~GiB for EGM-8B.  The 128-row measurements show the same
pattern (Appendix~\ref{app:cost-audit}).

\subsubsection{Specialist accuracy with LFPR applied}
\label{app:egm-lfpr-accuracy}

The cost audit above measures latency and memory only. Table~\ref{tab:egm-lfpr-accuracy}
scores accuracy on the same mechanism (LFPR's crop/guard/fusion pipeline
applied to each specialist's own predicted box) on the full 30,969-row
transfer split, using an already-archived per-row artifact with no new
inference: a paired image-cluster bootstrap (10,000 iterations, shared
draws across metrics).

\begin{table}[t]
\centering
\small
\begin{tabular}{lrrrr}
\toprule
System & Acc@0.5 $\Delta$ & Acc@0.75 $\Delta$ & Acc@0.9 $\Delta$ & mAcc $\Delta$ \\
\midrule
EGM-4B & +0.284 & +0.455 & +1.569 & +0.785 \\
& $[0.121,0.449]$ & $[0.147,0.776]$ & $[1.044,2.099]$ & $[0.626,0.940]$ \\
EGM-8B & +0.378 & +1.230 & +6.716 & +2.401 \\
& $[0.227,0.537]$ & $[0.915,1.545]$ & $[6.010,7.406]$ & $[2.226,2.577]$ \\
\bottomrule
\end{tabular}
\caption{Accuracy delta from applying LFPR's crop/guard/fusion mechanism to
each specialist's own predicted box, full 30,969-row transfer split, 95\%
paired image-cluster bootstrap intervals below each point estimate. Every
interval excludes zero: LFPR improves both specialists on every reported
metric, most sharply EGM-8B's comparatively weak Acc@0.9 ($+6.716$ points).
Absolute values appear in Table~\ref{tab:specialists} in the main text.}
\label{tab:egm-lfpr-accuracy}
\end{table}

These measurements close the marginal cost question for adding LFPR to each
specialist.  To reduce the hardware mismatch in a cross-system comparison, we
also reran the generalist baseline and full LFPR on the exact 128-row prefix on
the same RTX~3090 host.  The baseline uses 1.000 calls, 17.5 generated tokens,
and 1.041~s per row; LFPR uses 2.008 calls, 35.3 tokens, and 1.948~s.  Thus LFPR
is $1.871\times$ slower and uses $2.011\times$ the generated tokens, with peak
allocated memory increasing from 16.49 to 17.58~GiB.  On the same rows,
EGM-4B and EGM-8B alone are respectively $4.61\times$ and $4.37\times$ slower
than the generalist baseline; with LFPR they are $4.75\times$ and $4.68\times$
slower than generalist LFPR, and generate approximately $5.6\times$ as many
tokens.

This anchor is deliberately small and contains only one routed row.  The
generalist uses three timed repeats while EGM uses one, and the EGM shards ran
on different physical 3090s of the same SKU.  The comparison therefore places
the observed inference burden on one host; it does not replace the corrected
full-size generalist audit or establish a hardware-independent energy ranking.

\subsection{Specialist scaling and firewall detail}

This subsection was compressed in the main text's specialist-comparison
discussion (\S\ref{sec:specialists}) to save space; both facts below are
cited from there.  Against the frozen incumbent, the 8B specialist gains
$+2.099$ points at Acc@0.5 yet loses $7.168$ at Acc@0.9 and more paired rows
than it wins on IoU (11,428 against 19,195) -- a single permissive-threshold
number would rank this model first and its localization quality last.  The
4B$\to$8B scaling contrast reported in the main text carries 95\% CIs
$[0.171,1.011]$ at Acc@0.5 and $[-8.816,-6.730]$ at Acc@0.9.  On the firewall
audit (17,978 specialist-training images against 3,982 evaluation images,
zero overlap), the Acc@0.5 advantage's benchmark-specificity is concrete:
79.60\% against 87.05\% for the matched direct prompt on long expressions.
A coordinate-granularity check rules out quantization as an explanation
(Appendix~\ref{app:e3-transfer}).

\subsection{Historical selection and stopping rules}
\label{sec:prereg}

The historical workflow advanced a candidate when a one-sided 95\% lower
bound on Acc@0.5 exceeded zero and treated recomputation on reused rows as
motivating evidence.  The surviving records document at least fourteen named
candidate routes, in addition to resolution, prompt, guard, and fusion choices.
These rules reduced compute but were not registered in an external,
timestamped archive and did not adjust for selection across the full family.
Consequently, historical one-sided lower bounds are retained only to explain
which experiments were run; they are not confirmatory tests.  The primary
descriptive contrast in this manuscript is full LFPR minus the frozen
incumbent on the same 31,921 rows.  Acc@0.75, Acc@0.9, mAcc, source effects,
and all pilot results are exploratory secondary analyses.
\subsection{Exploratory source-stratified effects}

The source-restricted point differences are positive on both subsets: +0.875
points for COCO and +1.720 for Objects365.  One-sided lower bounds
are +0.693 and +1.260, respectively, but no interaction interval for the
difference between source effects was archived.  We therefore do not claim
heterogeneity or ``source consistency.''  The method uses no source-specific
threshold or prompt.  The router flagged 6,641 of 31,921 rows (20.8\%) and
agreed with the annotation-derived bucket on 89.83\% of rows; this is a
descriptive diagnostic, not a training or selection target.

\subsection{Pilot-to-full-split shift}

The pilot's own source-stratified effects (+1.30 points on its Objects365
rows) do not arrive in the same 62.3\%/37.7\% COCO/Objects365 mixture as
the official split, so a naive pilot-to-population extrapolation would
overstate the effect if it simply reused the pilot's own composition.
Re-weighting the pilot's per-source effects by the official split's
composition instead projects the +0.85-point pilot effect to approximately
+0.74 points, already below the +0.97-point gap to the working 89.5\%
comparison before the complete split is scored.  The full-split difference
turned out larger, not smaller, at +1.194 points once the resolution-routing
step (initially executed against the unrouted incumbent by implementation
error, corrected and re-run) was actually included.  This shift, in both
directions across the two corrections this project made to it, illustrates
why a pilot projection should not become an absolute ranking claim.  Because
both datasets belong to
the same adapted research program, the full-split interval is descriptive
rather than an independent prospective confirmation.

\subsection{Comparison to published Ref-L4 results}
\label{sec:leaderboard}

Table~\ref{tab:leaderboard} places the measured incumbent and LFPR against
published Ref-L4 test-split numbers collected from a single technical
report so that every external entry shares one leaderboard's reconciled
definitions \citep{glm45v2025}.  The frozen Qwen3-VL-8B incumbent
(88.531\%) already exceeds two published 7--9B-parameter systems and trails
the strongest published entry, a substantially larger mixture-of-experts
model, by 0.97 points.  LFPR's official result (89.725\%) exceeds that
entry by 0.225 points without changing the backbone.  This table is provided for
context and is not the basis of any claim in this paper: the published
figures come from a different evaluation protocol (prompt, parser, and in
one case an asterisked benchmark condition), so a comparison across rows
is cross-protocol and parameter-asymmetric, and Section~\ref{sec:sota}
explains why we do not convert Table~\ref{tab:leaderboard} into a ranking
statement.

\begin{table}[t]
\centering
\footnotesize
\setlength{\tabcolsep}{3pt}
\begin{tabular}{lrr}
\toprule
System & Params. & Acc@0.5 \\
\midrule
\textbf{Qwen3-VL-8B + LFPR (ours)} & \textbf{8B} & \textbf{89.73} \\
GLM-4.5V \citep{glm45v2025} & 106B (MoE) & 89.5 \\
GLM-4.6V \citep{glm45v2025} & -- & 88.9 \\
Qwen3-VL-8B incumbent (ours) & 8B & 88.53 \\
PLaMo 2.1-8B-VL \citep{glm45v2025} & 8B & 86.8 \\
GLM-4.1V-9B-Thinking \citep{glm45v2025} & 9B & 86.8 \\
Qwen2.5-VL-7B \citep{qwen25vl2025} & 7B & 80.8 \\
\bottomrule
\end{tabular}
\caption{Published Ref-L4 test-split Acc@0.5, collected from one
leaderboard table for a shared reference point, against our two directly
measured rows.  Rows are not evaluated under one identical protocol; see
text.}
\label{tab:leaderboard}
\end{table}

\subsection{Where the headroom comes from: a source- and size-stratified diagnosis}
\label{sec:diagnosis}

Before describing how the resolution floor and crop context were chosen,
we report the diagnostic that motivated targeting them at all.  Splitting
the 31,921-row test split by the annotation-derived target size gives a
strictly monotone accuracy gap: small targets score 78.92\% (6,665 rows,
20.9\% of the split), medium targets 90.15\% (16,311 rows, 51.1\%), and
large targets 92.73\% (8,945 rows, 28.0\%) -- a 13.81-point gap between the
smallest and largest bucket.  This association motivated the resolution
intervention, but target size is confounded with category, scene density,
expression content, and annotation properties and therefore does not identify
a resolution deficit by itself.

\begin{figure}[t]
\centering
\includegraphics[width=\linewidth]{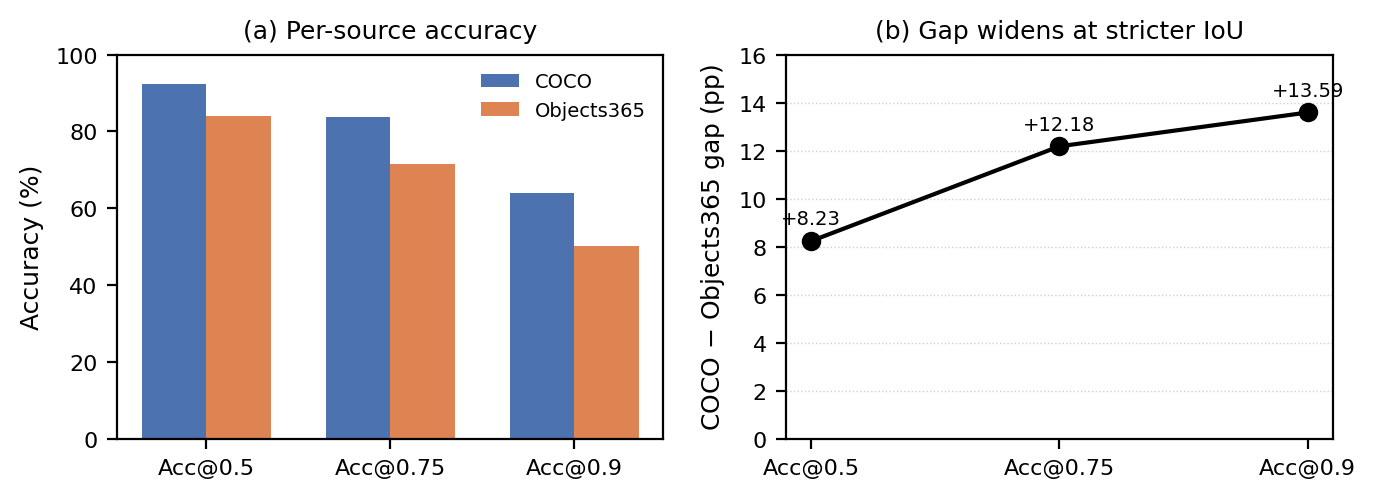}
\caption{Source-stratified accuracy on the recombined predictions.  The
COCO--Objects365 gap widens as the IoU threshold tightens (+8.23, +12.18,
+13.59 points), an association compatible with several localization and
domain-shift explanations.}
\label{fig:source-gap}
\end{figure}

A second, source-based split shows a different and complementary pattern.
Table~\ref{tab:source-diagnosis} and Figure~\ref{fig:source-gap} report
Acc@0.5, Acc@0.75, Acc@0.9, and mAcc separately for the COCO and
Objects365 rows of the same predictions.
The gap is present at every threshold and \emph{widens} as the threshold
becomes stricter (+8.23, +12.18, +13.59 points at Acc@0.5, Acc@0.75, and
Acc@0.9), a pattern consistent with a localization-precision difference.
The gap is also present within every size bucket (small +5.46, medium +6.25,
large +7.69 points), so target-size composition alone does not explain it.
These descriptive summaries cannot rule out source-specific category mix,
crowding, expression complexity, semantic errors, or annotation policy.  They
motivate, but do not causally validate, the resolution and crop components.

\begin{table}[t]
\centering
\footnotesize
\setlength{\tabcolsep}{3pt}
\begin{tabular}{lrrrr}
\toprule
Source & Acc@0.5 & Acc@0.75 & Acc@0.9 & mAcc \\
\midrule
COCO (62.3\%) & 92.32 & 83.76 & 63.87 & 78.81 \\
Objects365 (37.7\%) & 84.09 & 71.58 & 50.28 & 67.54 \\
\midrule
Gap & +8.23 & +12.18 & +13.59 & +11.27 \\
\bottomrule
\end{tabular}
\caption{Source-stratified accuracy on the recombined predictions
(post-resolution-diagnostic).  The gap widens at stricter thresholds and is
present within every size bucket, showing that target-size composition alone
does not account for the source association.}
\label{tab:source-diagnosis}
\end{table}

\subsection{Selecting and validating the resolution floor}
\label{sec:floor-selection}

The minimum pixel budget used in the router (Section~\ref{sec:method}) was
chosen with a small staged pilot on the small-target bucket before
committing to a full run, following the historical rule fixed at pilot
time that a pilot without a positive signal does not earn a full-scale
run.  On an
800-row small-bucket sample, a moderate floor ($\sim2.7\times$ the default)
gave +0.75 points with a one-sided LCB of $-0.99$ (not distinguishable from
zero); the floor used throughout this paper ($4{,}194{,}304$ pixels,
$\sim10.6\times$ the default) gave +1.875 points with LCB $+0.25$, clearing
zero on the pilot sample and motivating the full run.

Those pilots, and the full Ref-L4 run they authorised, were later found to
have requested the floor through a processor keyword this checkpoint discards
(Section~\ref{sec:transfer}), so they measured default-resolution work.  We
therefore re-ran the routed pass on all 6,641 flagged Ref-L4 rows with the
override verified in effect.  It changes the predicted box on 6,591 of them
(99.2\%), against 627 rows of $\pm1$-unit numerical jitter before.  Applying
the floor to the routed rows and leaving every other row untouched moves the
full 31,921-row split from 88.531\% to 89.230\% at Acc@0.5 ($+0.699$), from
77.410\% to 79.321\% at Acc@0.75 ($+1.911$), from 55.778\% to 59.030\% at
Acc@0.9 ($+3.252$), and mean IoU from 80.936\% to 81.949\%.  This
routing-only measurement was itself later superseded: the crop stage had a
second, independent bug at the time (it was cropping the unrouted baseline
rather than the routed prediction), so the 89.230\% figure and its 80.936\%
baseline predate that fix.  The canonical full-policy numbers, with both
bugs fixed, are Table~\ref{tab:main-results}'s 88.531\%$\to$89.725\% at
Acc@0.5 and 80.944\%$\to$82.648\% mean IoU (Appendix~\ref{app:refl4-meaniou-erratum}); readers should treat
Table~\ref{tab:main-results} as the canonical baseline and the figures in
this paragraph as a superseded intermediate diagnostic of the routing
stage alone.

The direction of the original claim survives; its magnitude and its mechanism
do not.  The gain is larger than the withdrawn $+3.30$-point small-bucket
figure and, more importantly, it is not confined to the smallest targets: the
medium bucket gains $+16.51$ points at Acc@0.9 while the large bucket gains
$+1.88$.  The earlier conclusion that the floor does not generalize past the
small bucket was an artifact of the same defect, since the medium-bucket
extension pilot that produced $-0.125$ points also ran at default resolution
and was therefore comparing the incumbent against itself.  The router's
restriction to the small bucket is consequently a conservative operating
choice rather than an empirically forced one, and widening it is a concrete
avenue this work leaves open.

\subsection{Exploratory resolution checks do not establish a peak}
\label{sec:headroom-exhausted}

Two prompt variants were piloted against the frozen default prompt.  A
\emph{tight-box} instruction changed Acc@0.5 by +0.125 points (historical
one-sided LCB $-1.13$), and a \emph{fine-grained} instruction changed it by
+0.25 points (LCB $-1.00$).  These are exploratory null results.

The archived resolution summaries reveal a reference inconsistency in the
earlier ladder figure, which is therefore removed.  On an 800-row small-target
sample, the $1{,}048{,}576$- and $4{,}194{,}304$-pixel summaries report
default-relative changes of +0.750 and +1.875 points.  The
$8{,}388{,}608$-pixel summary instead reports +1.125 points relative to the
$4{,}194{,}304$ arm on the same 800 rows (82.875\% versus 81.750\%).  A
stepwise increment cannot be plotted as a default-relative ordinate or used
to conclude that the curve peaked.  Until the row-level predictions are
rescored against one common reference, the defensible conclusion is only that
$4{,}194{,}304$ was the historically selected operating point; remaining
resolution headroom is unresolved.

\subsection{An unguarded self-correction baseline regresses the headline metric}
\label{sec:self-correction-baseline}

Section~\ref{sec:related} distinguishes LFPR from iterative visual
self-correction methods by its single guarded attempt.  To make this
distinction an empirical rather than only a design point, we ran a
matched, zero-shot self-correction control on an 800-row pilot: the same
frozen incumbent is shown the original image with its own first box drawn
as an overlay and asked to return one replacement box, with no guard and
no fusion.  This regresses Acc@0.5 by $-2.00$ points (one-sided LCB
$-3.14$) and Acc@0.75 by $-2.125$ points (LCB $-3.78$); only Acc@0.9
improves (+2.25 points, LCB $+0.12$), and mAcc regresses by $-1.06$ points
(LCB $-2.14$).  An ungated second opinion is therefore not a substitute for
LFPR's guarded crop: it can occasionally sharpen an already-close box
(the Acc@0.9 effect) while making the incumbent's committed answer worse
on average, which is the same asymmetry the self-correction mirage
literature reports for oracle-free stopping \citep{tripathy2026itervisual}.
LFPR's guards are best understood as an explicit, non-learned stopping
rule that keeps this same mechanism from being applied unconditionally.

\subsection{Matched negative controls define the claim}
\label{sec:negative-controls}

The positive full LFPR result is interpretable because nearby alternatives were
not silently promoted.  A supervised coarse-to-fine adaptation of the
incumbent, fine-tuned for 10,000 steps on external Objects365 boxes and
evaluated on a reserved, never-previously-read 3,314-row confirmation
split through the same frozen evaluation harness, decreased Acc@0.5
by 0.543 points, with LCB -1.264: adapting the weights directly, using the
same external data source LFPR's own diagnostics draw on, did not transfer
to a gain.  A released third-party checkpoint built for parallel box
decoding was evaluated zero-shot on the shared pilot with zero
runtime errors and decreased the metric by 15.80 points; this checkpoint's
own training-image overlap with Ref-L4 was never resolved by its authors,
so it was treated as diagnostic-only regardless of its result.  SAM3
boundary projection with the same midpoint
fusion produced only +0.10 points on a fresh pilot, with LCB
-0.20; a retrospective check that applied the same fusion rule to the
projection pilot's own
pilot rows had suggested +0.30 points, but this signal did not survive a
prospective confirmation on a disjoint holdout, illustrating why this paper
treats retrospective checks as motivating evidence only, never as
confirmation.  Full crop replacement and standalone conservative
fusion improved strict-IoU metrics but did not establish the required
headline gain on their own.  Finally, scaling the incumbent within its own
model family to a $4\times$-larger 32B checkpoint, under the identical
frozen prompt, parser, and evaluator, also closed negative: +0.26 points at
Acc@0.5 with LCB $-2.45$, a regression at Acc@0.9 ($-1.32$ points, LCB
$-4.56$), a failing effect on the Objects365 subset ($-0.53$ points), and a
COCO LCB ($-2.07$) outside the historical $-0.5$-point band.  These controls
show that several nearby alternatives failed under their respective pilots;
because populations differ and a full same-row factorial table is absent,
they do not isolate the causal contribution of routing, guards, or fusion.

\section{Additional Metric Definitions}
\label{app:metrics}

For an image $j$ carrying expressions $i\in\mathcal{I}_j$, the paired
image-level bootstrap samples images rather than individual expressions.
Let $m_j^*$ be the multiplicity of image $j$ in one bootstrap draw.  The
treatment effect is

\begin{equation}
  \widehat{\Delta}_t^*=
  \frac{\sum_j m_j^*\sum_{i\in\mathcal{I}_j}
  [u_t(\widehat b_i,y_i)-u_t(b_{0i},y_i)]}
  {\sum_j m_j^*|\mathcal{I}_j|}.
\end{equation}

The reported percentage-point effect is $100\widehat{\Delta}_t$.  This is an
expression-weighted estimand with image-cluster resampling.  mAcc is
computed before taking the treatment difference so that each threshold has
equal weight.  Source effects use the same definition after restricting
$i$ to the source subset.  No multiplicity correction was used historically;
Acc@0.75, Acc@0.9, mAcc, source slices, and pilot comparisons are therefore
reported as an exploratory family rather than confirmatory tests.

\end{document}